\documentclass[conference]{IEEEtran}
\usepackage{cite}
\usepackage{amsmath,amssymb,amsfonts}
\usepackage{graphicx}
\usepackage{textcomp}
\usepackage{xcolor}

\usepackage{booktabs}
\usepackage{algorithm}
\usepackage{algpseudocode}
\usepackage{multirow}
\usepackage{colortbl}
\definecolor{lightgray}{gray}{0.9}
\usepackage{hyperref}
\usepackage{cleveref}

\crefname{equation}{Equation}{Equations}
\Crefname{equation}{Equation}{Equations}
\crefname{table}{Table}{Tables}
\Crefname{table}{Table}{Tables}
\crefname{figure}{Fig.}{Figs.}
\Crefname{figure}{Fig.}{Figs.}
\crefname{section}{Section}{Sections}
\Crefname{section}{Section}{Sections}
\crefname{algorithm}{Algorithm}{Algorithms}
\Crefname{algorithm}{Algorithm}{Algorithms}

\def\eg{\textit{e.g.}}
\def\ie{\textit{i.e.}}
\def\etal{\textit{et al.}}

\def\BibTeX{{\rm B\kern-.05em{\sc i\kern-.025em b}\kern-.08em
    T\kern-.1667em\lower.7ex\hbox{E}\kern-.125emX}}
\begin{document}

\title{Test-time Adaptation Meets \\Image Enhancement: Improving Accuracy via \\Uncertainty-aware Logit Switching}

\author{\IEEEauthorblockN{Shohei Enomoto\IEEEauthorrefmark{1}, Naoya Hasegawa\IEEEauthorrefmark{2}, Kazuki Adachi\IEEEauthorrefmark{1}, Taku Sasaki\IEEEauthorrefmark{1},\\
Shin'ya Yamaguchi\IEEEauthorrefmark{1}\IEEEauthorrefmark{3}, Satoshi Suzuki\IEEEauthorrefmark{1} and Takeharu Eda\IEEEauthorrefmark{1}}
\IEEEauthorblockA{
\IEEEauthorrefmark{1}NTT, Tokyo, Japan 
\IEEEauthorrefmark{2}University of Tokyo, Tokyo, Japan 
\IEEEauthorrefmark{3}Kyoto University, Kyoto, Japan\\
\IEEEauthorrefmark{1}\{shohei.enomoto, kazuki.adachi, taku.sasaki, shinya.yamaguchi, satoshixv.suzuki, takeharu.eda\}@ntt.com \\
\IEEEauthorrefmark{2}hasegawa-naoya410@g.ecc.u-tokyo.ac.jp}

}


\maketitle

\begin{abstract}
Deep neural networks have achieved remarkable success in a variety of computer vision applications.
However, there is a problem of degrading accuracy when the data distribution shifts between training and testing.
As a solution of this problem, Test-time Adaptation~(TTA) has been well studied because of its practicality.
Although TTA methods increase accuracy under distribution shift by updating the model at test time, using high-uncertainty predictions is known to degrade accuracy.
Since the input image is the root of the distribution shift,  we incorporate a new perspective on enhancing the input image into TTA methods to reduce the prediction's uncertainty.
We hypothesize that enhancing the input image reduces prediction's uncertainty and increase the accuracy of TTA methods.
On the basis of our hypothesis, we propose a novel method: Test-time Enhancer and Classifier Adaptation~(TECA).
In TECA, the classification model is combined with the image enhancement model that transforms input images into recognition-friendly ones, and these models are updated by existing TTA methods.
Furthermore, we found that the prediction from the enhanced image does not always have lower uncertainty than the prediction from the original image.
Thus, we propose logit switching, which compares the uncertainty measure of these predictions and outputs the lower one.
In our experiments, we evaluate TECA with various TTA methods and show that TECA reduces prediction's uncertainty and increases accuracy of TTA methods despite having no hyperparameters and little parameter overhead.
\end{abstract}


\section{Introduction}
\label{sec:intro}
Deep neural networks (DNNs) have achieved remarkable success in various computer vision applications, such as classification, segmentation, and object detection. 
However, DNNs have a problem with accuracy degradation when the data distribution shifts between training and testing.
This problem occurs frequently in outdoor environments, such as autonomous driving and smart cities, due to changes in weather and brightness~\cite{diamond2021dirty,in-c,Zendel_2018_ECCV}.
To solve this problem, DNNs need to be more robust to these distribution shifts.

One of the most practical settings to improve the robustness of DNNs is Test-time Adaptation~(TTA).
In the TTA setting, a pre-trained model is given, and the model is tested and updated simultaneously using unlabeled target data.
The model maintains accuracy by adapting to distribution shifts during testing.
The TTA setting has the advantages such as working under realistic distribution shifts~\cite{cotta,rotta,fedthe} and privacy protection because it does not store data.
This setting has attracted the attention of many researchers because of its generality, flexibility, and practicality.

The typical TTA method uses predictions to update a model~\cite{tent}.
Under continuously changing target distributions, the prediction becomes highly uncertain and noisy, leading to accuracy degradation~\cite{eata}.
To address this degradation, existing studies have proposed uncertainty reduction through an improved TTA framework~\cite{cotta} and an improved loss function without uncertain predictions~\cite{eata}.
Although these methods have increased accuracy by eliminating the adverse effects of uncertain predictions, a large gap remains between the accuracy of the source and target distributions.
This is because these methods use original input images, which are the root of the distribution shift, without any improvements.
Therefore, we focus on improving image quality, while existing TTA studies have focused on improving the framework and loss.
We hypothesize that improving image quality can reduce the uncertainty of the predictions and that updating the model by using such images further increases the accuracy of TTA methods.

To generate high-quality images, we use an image enhancement model~\cite{urie,task,when,effect,lp,resize,dyntta,bts}, which transforms distorted images into ones that can be easily recognized by classification models.
Since enhanced images are known to increase the accuracy of classification models, we hypothesize that they decrease the uncertainty of the predictions and are suitable for TTA methods.
On the basis of our hypothesis, we propose a novel method, Test-time Enhancer and Classifier Adaptation~(TECA), which combines an image enhancement model and a classification model into a single one and updates it in the TTA manner. 
By updating the unified model during testing, TECA provides predictions with low-uncertainty even under unknown shifted distributions, allowing stable updating of TTA methods.
In addition to combining two models, we present an important finding that the image enhancement model does not always reduce the prediction's uncertainty, indicating that updating classification models with enhanced images is not always appropriate.
To overcome this problem, we propose \textit{Logit Switching}~(LS).
LS compares the uncertainty of predictions from the original image and enhanced one, and outputs the one with the lower uncertainty
Furthermore, we introduce two modules: \textit{Synchronizing Parameter Updating Speed}~(SPUS), which balances the different speeds of the model update in LS and stabilizes the update, and \textit{Freezing Batch Normalization Statistics}~(FBNS), which preserves the source knowledge of an image enhancement model necessary to enhance the image quality.
By incorporating LS with SPUS and FBNS, TECA can effectively reduce the uncertainty of the classification predictions, and as a result, the TTA methods increase the accuracy much more.
TECA focuses on transforming the input images, which existing TTA methods do not deal with, and can be easily combined with existing TTA methods.

We conducted experiments with the continual TTA task~\cite{cotta} on the ImageNet dataset~\cite{in-c} and with the domain generalization benchmarks~\cite{domainbed}.
We show that combining state-of-the-art TTA methods with TECA reduces prediction's uncertainty and increases accuracy.
TECA shows higher training efficiency than simply increasing the number of the classification model parameters, despite its low parameter overhead.
We also show that TECA works well with a variety of classification model architectures.

The contributions of this paper are as follows:
\begin{itemize}
\item We propose a novel method, TECA, which updates the image enhancement and classification models at test time from predictions with low uncertainty.
TECA is simple yet effective, with no hyperparameters and low overhead, but improves the accuracy of the TTA methods.
\item We show in experiments that TECA decreases the error rate of TTA methods for various distribution shifts.
TECA is more parameter effective than simply increasing the number of the classification model parameters and works with a variety of classification models.
\end{itemize}

\section{Related Work}
\subsection{Test-time Adaptation}
Test-time Adaptation~(TTA) is a setting that updates the model under distribution shift online with each mini-batch, without accessing the source dataset or labeled target data.
This problem setting has attracted much attention because it is more practical and has fewer constraints than Domain Generalization and Domain Adaptation.
As pioneering work, Wang \etal~\cite{tent} proposed TTA setting.
They found that the entropy of predictions correlates with accuracy and proposed Tent, which updates a classification model through entropy minimization.
To further improve the performance of TTA methods, numerous studies were conducted, such as using data augmentation to make predictions consistent~\cite{memo,tta_augmix,contrastive}, introducing a teacher-student training framework~\cite{cotta,petal}, using meta-learning~\cite{conjugate,mt3}, and aligning intermediate representations with prototypes~\cite{ttaps,robustifying,cafa,cafe,t3a}.
Since models are continuously updated online in the TTA setting, reducing their computational cost~\cite{eata,ecotta} and avoiding catastrophic forgetting~\cite{eata,cotta,petal} are also important research problems.
In addition, the application of TTA to various recognition tasks~\cite{ev_tta,mm_tta,tta_denoise,tta_vdu,tta_video,test,auxadapt,tta_stain} and its extension to more realistic distribution shift problem settings~\cite{cotta,rotta,fedthe} has also attracted attention.

Although many TTA studies have been conducted, a large gap remains between the accuracy of the source and target distributions.
To close this gap, we incorporate a new perspective on enhancing the input image into TTA methods.
We use an image enhancement model to transform the input image into a recognition-friendly image, thus increasing the performance of TTA methods.

\subsection{Input Adaptation} 
Input adaptation improves recognition accuracy by transforming input images into ones that are easier to recognize, \ie, \textit{recognition-friendly}, images.
Here we introduce two approaches: image enhancement, an image transformation method based on DNNs; and visual prompt adaptation, which adds trainable noise to the input image.

\subsubsection{Image Enhancement} 
In image enhancement, the DNN, which we call the image enhancement model, learns image transformations that minimize losses of the recognition model.
Sharma \etal~\cite{filter} improves recognition accuracy by emphasizing image-specific details with a DNN-based enhancement filter.
The subsequent work~\cite{urie,task,when,effect,lp,resize,dyntta,bts} uses DNN to transform corrupted images into recognition-friendly ones.

We aim to incorporate these input adaptation methods into TTA methods to further increase accuracy by reducing prediction's uncertainty from enhanced images.
Updating an image enhancement model at test time has never been conducted and allows for predictions with low-uncertainty even under distribution shift.
Furthermore, to avoid harmful updates caused by high-uncertainty prediction,  we propose switching to the predictions with low-uncertainty ones to update the model.

\subsubsection{Visual Prompt Adaptation}
To improve the robustness of DNNs, several studies have been conducted to update the visual prompt~\cite{vp,vpt}, which is trainable noise, at test time~\cite{vpt_tta,decorate,vp_conv}.
Unlike the image enhancement model, the visual prompt does not use a DNN, but transforms the input image by adding the noise.
In the visual prompt adaptation, the recognition model is frozen and only the visual prompt is updated during testing, as well as in TTA methods.

Our study is similar to these studies in terms of combining the image enhancer and TTA methods, although it differs in that it simultaneously updates the image enhancement model and the classification model from the switched low-uncertainty predictions.

\section{Test-Time Enhancer and Classifier Adaptation~(TECA)}

\subsection{Problem Definition}
\label{sec:problem}
Given the classification model $f^{\mathrm{cls}}_{\theta_0}(x)$ and the image enhancement model $f^{\mathrm{enh}}_{\phi_0}(x)$ trained on the source dataset, our goal is to update these models simultaneously during testing and adapt them to the target domain to increase their accuracy.
We treat these two models as one model $f_{\Theta_0}$, where $\Theta_0 = [\theta_0, \phi_0]$, and apply TTA methods.
Provided with the mini-batch of target data $x^T_t$ at time $t$ sampled from the unlabeled target domain dataset, $x^T_t$ is fed into the model $f_{\Theta_t}$ and the model outputs a prediction $f_{\Theta_t}(x^T_t)$.
The model is then updated $\Theta_{t}\xrightarrow{}\Theta_{t+1}$ from $f_{\Theta_t}(x^T_t)$.
The loss used to update the model depends on the TTA method applied (\eg, Tent uses entropy minimization).

\cref{tab:problem_settings} lists the main differences between the TECA setting and the existing adaptation settings.
Neither setting requires access to the source dataset and target data labels, although each setting has different updated parameters and given components.
In the TTA setting, the classification model is updated at test time by using unlabeled target data.
The image enhancement and visual prompt adaptation settings transform the target image during testing but the given components are different.
In the image enhancement setting, an image enhancement model transforms the target image into a recognition-friendly one.
In the visual prompt adaptation setting, the visual prompt, the trainable noise that increases recognition accuracy, is updated during testing.
The TECA problem setting updates both the classification and image enhancement models at test time.

\begin{table}[t!]
\centering
\caption{
Main differences between the TECA problem setting and the existing adaptation problem setting.
Existing settings maintain classification accuracy under distribution shift by updating classification models or transforming target images at the test time.
The TECA setting updates parameters in both the classification and the image enhancement models at the test time.
$v_{\psi_0}$ is a pretrained visual prompt.
}
\begin{tabular}{lcccc}
\toprule
 Setting & Updated & Given    \\ \midrule
    TTA & $\theta_t$& $f^{\mathrm{cls}}_{\theta_0}$   \\ 
    Image Enhancement & -- & $f^{\mathrm{cls}}_{\theta_0}$, $f^{\mathrm{enh}}_{\phi_0}$  \\ 
    Visual prompt adaptation & $v_{\psi_t}$ & $f^{\mathrm{cls}}_{\theta_0}$, $v_{\psi_0}$  \\ \midrule 
    TECA & $\theta_t$, $\phi_t$ & $f^{\mathrm{cls}}_{\theta_0}$, $f^{\mathrm{enh}}_{\phi_0}$  \\ \bottomrule
\end{tabular}
\label{tab:problem_settings}
\end{table}

\subsection{Analysis of Image Quality, Uncertainty, and Error Rate}
\begin{figure}[t]
\centering
\includegraphics[width=0.8\linewidth]{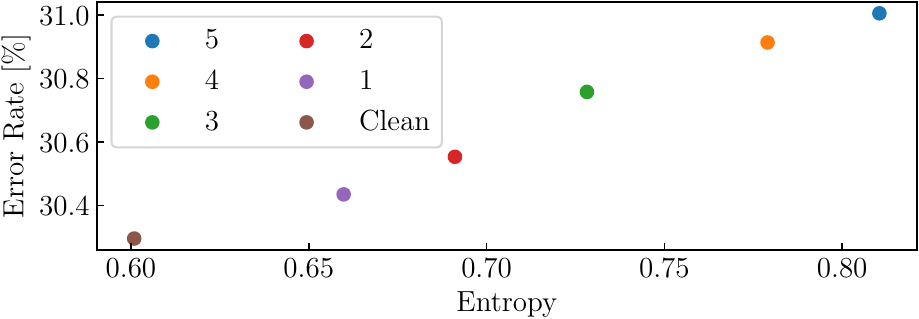}
\caption{
Preliminary experiments on our hypothesis.
We updated ResNeXt~\cite{resnext} with Tent on the CIFAR-100-C dataset and measured error rates and entropy.
We tested clean images and ones corrupted at five severity levels, where higher severity indicates stronger corruption and lower image quality, while lower severity indicates cleaner images and higher image quality.
}
\label{fig:preliminary}
\end{figure}

In this paper, we hypothesize that improving image quality can reduce the uncertainty of the predictions and that updating the model by using such images further increases the accuracy of TTA methods.
To verify our hypothesis, we analyzed the relationship between error rate and uncertainty in various image quality, as shown in \cref{fig:preliminary}.
We found that the error rate and entropy were inversely proportional to the image quality.
In other words, the higher the image quality, the smaller the error rate and entropy, and the lower the image quality, the larger the error rate and entropy.
This suggests that high-quality images improve the accuracy of the TTA method, which supports our hypothesis.

\subsection{Methodology}
\begin{figure*}[t]
\centering
\includegraphics[width=0.7\linewidth]{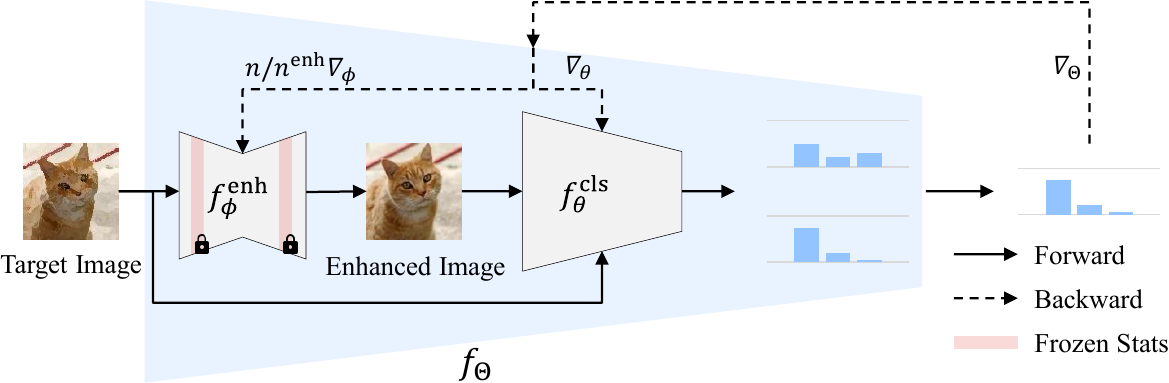}
\caption{
Overview of TECA.
The image enhancement model transforms the target image into a recognition-friendly one.
The classification model predicts both original and recognition-friendly images.
TECA compares the uncertainty of the predictions and switches predictions to the lower one.
The model is updated from the prediction using the TTA methods.
At this time, the gradient of the image enhancement model is rescaled to align the parameter update speed of each model that stabilizes the updates.
Additionally, the BN statistics of the image enhancement model are frozen to preserve source knowledge.
}
\label{fig:overview}
\end{figure*}

\begin{figure}[t]
\centering
\includegraphics[width=0.8\linewidth]{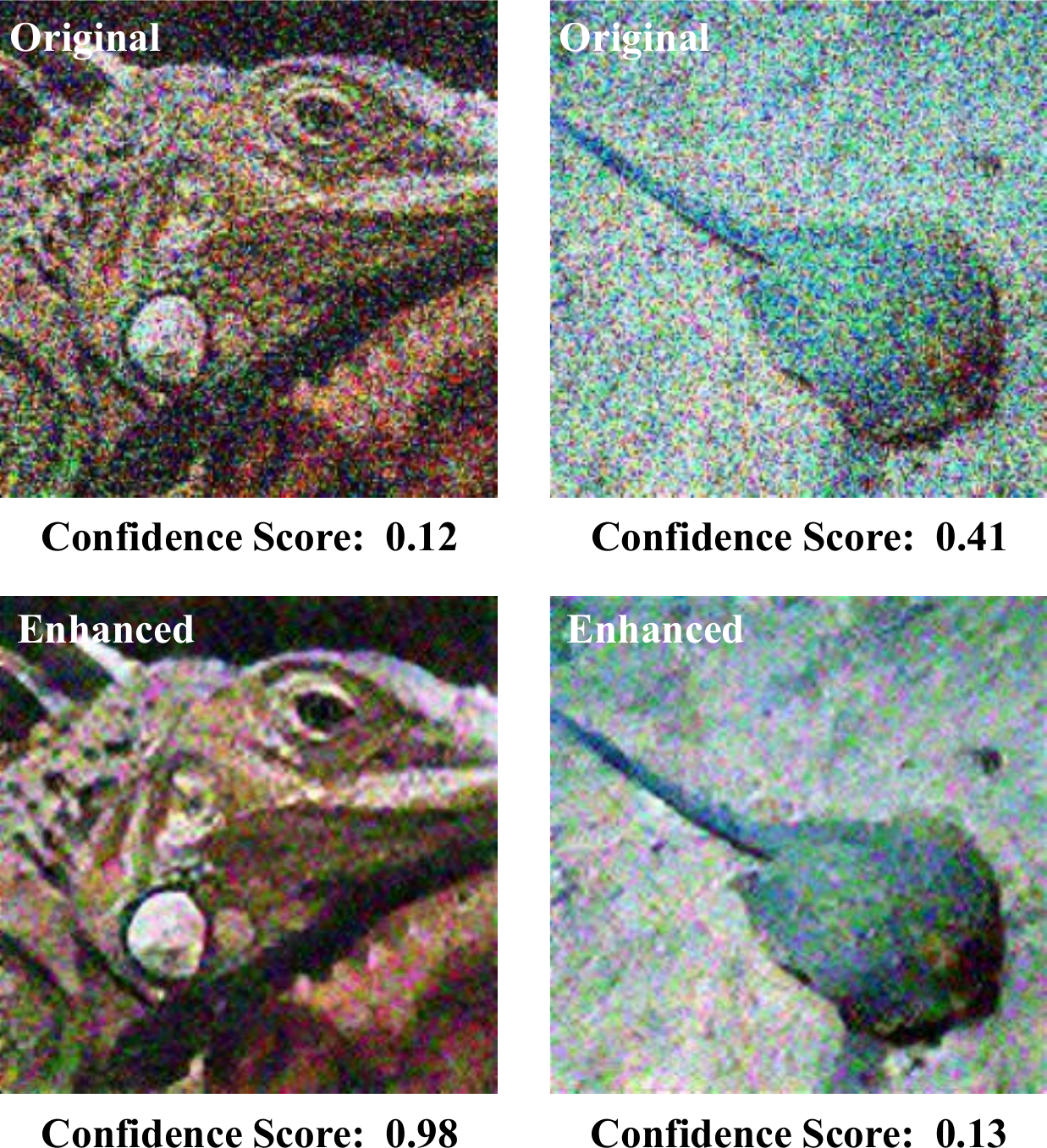}
\caption{
Gaussian noise at severity level 5 images and URIE~\cite{urie}-enhanced images and their confidence score in the ImageNet-C dataset.
The top and bottom rows are the original and enhanced images, respectively.
}
\label{fig:urie_output}
\end{figure}

\cref{fig:overview} and \cref{alg:teca} show the overview and pseudo-code of TECA.
To further reduce prediction's uncertainty and increase the accuracy of existing TTA methods, we study improvements in image quality.
\cref{fig:urie_output} shows the original and enhanced images and their confidence score.
We found that performing classification with enhanced images does not always achieve a higher confidence score, an uncertainty measure, than one with the original images.
The straightforward input adaptation approach uses only enhanced images, and thus it is not suitable in the TTA setting.
Therefore, we propose LS that compares the uncertainty of the predictions from the original and enhanced images and then outputs the lower one.
Furthermore, we introduce two modules to increase the stability of TECA: SPUS to align the parameter update speeds of the classification and image enhancement models, and FBNS to freeze the BN statistics of the image enhancement model to retain source knowledge for recognition-friendly image transformation.

\begin{algorithm}[t]
\caption{Test-Time Enhancer and Classifier Adaptation~(TECA)}
\label{alg:teca}
\textbf{Initialization:} Model $f_{\Theta_0}$ composed of source pre-trained classification model $f^{\mathrm{cls}}_{\theta_0}$ and image enhancement model $f^{\mathrm{enh}}_{\phi_0}$. \\
\textbf{Input:} For each time step $t$, current stream of data $x_t$. 
\begin{algorithmic}[1]
\State Get the enhanced image $f^{\mathrm{enh}}_{\phi_t}(x_t)$.
\State Calculate predictions $f^{\mathrm{cls}}_{\theta_t}(x_t)$ and $f^{\mathrm{cls}}_{\theta_t}(f^{\mathrm{enh}}_{\phi_t}(x_t))$.
\State Compare the confidence score of predictions and select the higher one as $f_{\Theta_t}(x_t)$ by \cref{eq:high_q_out}.
\State Calculate gradients $\nabla_\Theta$ by TTA loss.
\State Rescale gradients $\nabla_\phi$ by \cref{eq:spus}.
\State Update $\Theta_t$ using gradient descent.
\end{algorithmic}
\textbf{Output:} Prediction $f_{\Theta_t}(x_t)$; Updated model $f_{\Theta_{t+1}}$.
\end{algorithm}

Given an image enhancement model, TECA can be easily combined with existing TTA methods to further increase accuracy, since TECA has no hyperparameters.
We describe the details of our methodology in the following paragraphs.

\paragraph{Given Models.}
In the TECA setting, we assume that a classification model and an image enhancement model that are pre-trained on the source dataset are given as stated in~\cref{sec:problem}.
The classification model does not require a special process for the TECA setting.
In this paper, we use URIE~\cite{urie} as an image enhancement model.
URIE has a U-Net~\cite{u-net} type DNN structure that improves the robustness of the classification model with a few trainable parameters.
URIE is trained to minimize the loss of the frozen pre-trained classification model on the source dataset with artificial distribution shifts~(\eg, data augmentation).
We refer to this frozen pre-trained classification model as \textit{enhancer partner}.

\paragraph{Logit Switching (LS).}
As shown in \cref{fig:urie_output}, the image enhancement model does not always reduce prediction's uncertainty.
Adapting models from uncertain predictions should be avoided because it leads to error accumulation and catastrophic forgetting~\cite{eata}.
Therefore, to solve this problem, we propose LS that obtains predictions from both the enhanced and original images, compares the predictions with confidence scores, and then switches the prediction to the higher confidence one and uses it for model updating.
The LS is shown in \cref{eq:high_q_out}.

\begin{align}
\label{eq:high_q_out}
f_{\Theta_t}(x) &= \left\{ \begin{array}{ll}
f^{\mathrm{cls}}_{\theta_t}(f^{\mathrm{enh}}_{\phi_t}(x)) \; & (\text{if ${\max({p}}^{\mathrm{enh}})$ \textgreater ${\max({p}}^{\mathrm{cls}})$)} \\
f^{\mathrm{cls}}_{\theta_t}(x) \; & (\text{else}), \\
\end{array} \right.
\end{align}
where $x$ is an input image and $p^{\mathrm{enh}}$ and $p^{\mathrm{cls}}$ are the softmax probability predictions from the enhanced and original images, respectively.

\paragraph{Synchronizing Parameter Updating Speed (SPUS).}
Whereas the classification model $f^{\mathrm{cls}}_{\theta}$ is updated from all predictions, the image enhancement model $f^{\mathrm{enh}}_{\phi}$ is not updated if the predictions from the original images are selected because there is no backward path.
This causes a difference in the updated speed, which harms the accuracy~\cite{fast_advprop}.
To prevent this, inspired by \cite{fast_advprop}, we introduce SPUS, which rescales the gradient and ensures a similar parameter updating speed.
Let $n$ be the batch size at the test time, $n^{\mathrm{enh}}$ is the number of enhanced images employed in \cref{eq:high_q_out}, and we rescale the gradient of the image enhancement model by

\begin{align}
\label{eq:spus}
\nabla_\phi \leftarrow (n/n^{\mathrm{enh}}) \nabla_\phi .
\end{align}

\paragraph{Freezing BN Statistics (FBNS).}
\textit{BN Adapt}~\cite{bn_adapt} only updates the BN statistics from input data at the test time.
Since it is a simple and effective TTA method, many TTA methods have built-in BN Adapt.
However, we experimentally found that BN Adapt is not suitable for the image enhancement model and decreases the accuracy (see \cref{sec:param_range} for details).
We solve this problem by freezing the BN statistics of the image enhancement model during testing.

\section{Experiments}
\label{sec:exp}
We combined TECA with the several TTA methods and evaluated them in the standard continual TTA (CTTA)~\cite{cotta} task and domain generalization benchmarks.
We show that TECA increases accuracy of existing TTA methods despite the small overhead and works with a variety of classification models.

\subsection{Experimental Setup}

\paragraph{Dataset and Tasks.}
For the standard CTTA task, we conducted experiments on the ImageNet-C dataset~\cite{in-c} to evaluate the robustness against common image corruption.
This dataset contains 15 types of corruptions with 5 levels of severity.
Following existing studies, we used the largest corruption severity level 5 and 5,000 samples were extracted from the test set by RobustBench to evaluate the error rate in each corruption.
For the domain generalization benchmarks, we experimented with four datasets, VLCS, PACS, OfficeHome, and TerraIncognita, using the DomainBed library~\cite{domainbed}.

We experimented with a classification model pre-trained on the ImageNet1K dataset published by torchvision (excluding DeiT~\cite{deit}), with TTA methods.
We evaluated the avoidance of error accumulation and catastrophic forgetting by TTA methods in the CTTA task, where the distribution is continuously changing.
For the standard CTTA task, we evaluated TTA methods on a corrupted test set and then evaluate it on the next corrupted test set without resetting the model and optimizer.

\paragraph{TTA Methods.}
We refer to the method that uses only the classification model as \textit{Source} and the method that uses the enhanced image by URIE as input to the classification model as \textit{Input Adapt}.
In these methods, the models are frozen and not adapted at test time.
For the CTTA task, we used four TTA methods.
BN Adapt~\cite{bn_adapt} updates only BN statistics at the test time. 
Note that when we combine TECA with BN Adapt, we do not freeze BN statistics.
\textit{Tent}~\cite{tent} updates only the BN layer through entropy minimization.
\textit{EATA}~\cite{eata} extends Tent by efficient training with active sample selection and avoiding catastrophic forgetting with Elastic Weight Consolidation.
\textit{CoTTA}~\cite{cotta} uses a self-training framework, augmentation-averaged pseudo-labels, and stochastic restoration to deal with error accumulation and catastrophic forgetting in continually changing distributions.
For the domain generalization benchmarks, we used \textit{T3A}~\cite{t3a}, a TTA method which adjusts the linear classifier (the last layer of DNNs) at test-time. 

\paragraph{Implementation Details.}
For the CTTA task, we used SGD with momentum 0.9, batch size 64, learning rate of 0.01 for CoTTA and RoTTA, and 0.00025 for Tent and EATA.
For the domain generalization benchmarks, we used batch size 32.
The other hyperparameters for each method and task follow the original setting.

The original training setup for URIE uses the corruption dataset and is not suitable for the experimental setup for the CTTA task, which is evaluated on the corruption dataset.
Therefore, we trained URIE on the augmented dataset in accordance with the blind setting proposed in~\cite{dyntta}. 
For the CTTA task, we used AugMix~\cite{augmix} and DeepAugment~\cite{in-r}.
As with the AugMix paper, data augmentations used by AugMix\footnote{autocontrast, equalize, posterize, rotate, solarize, shear x, shear y, translate x, translate y} do not include the test corruption in the ImageNet-C dataset.
For the domain generalization benchmarks, we used AugMix with four additional operations\footnote{color, contrast, brightness, sharpness} because there are no data augmentations that overlap with the test set.
In this setting, URIE learns to enhance the features needed to classify out-of-distribution data from diverse data generated by the data augmentation.
This enables URIE to increase classification accuracy for unknown distribution shifts, even though it does not learn the distribution shifts at test time.
We used ResNet-50 as the enhancer partner when training URIE, except for the experiments in \cref{sec:other}.

\subsection{Experimental Results}
\subsubsection{Results of the CTTA Task}
\cref{tab:standard} shows the results of the standard CTTA task.
Combining TTA methods with TECA reduces error rates for almost all corruptions except for a few and on average improves error rates for all methods.
The combination of Tent and TECA outperforms the state-of-the-art TTA methods, EATA and CoTTA, indicating that adaptation efficiency with TECA is high.
This result indicates that enhancing the input image, which is the root of the distribution shift, is more beneficial for TTA methods than improving the framework or loss.
The combination of CoTTA and TECA shows the best overall result, reducing the average error rate by 3.89 points compared with CoTTA alone.

\begin{table*}[t]
\centering
\caption{
Classification error rate of ResNet-50 for the standard ImageNet CTTA task. 
Mean indicates the average error rate for all corruption test sets. 
The best and second results are \textbf{bolded} and \underline{underlined}.
}
\resizebox{\textwidth}{!}{
\begin{tabular}{l|c|lllllllllllllll|c}\toprule
\multicolumn{2}{l|}{Time} & \multicolumn{15}{l|}{$t\xrightarrow{\hspace*{15.0cm}}$}& \\ \midrule
Method                          & \rotatebox[origin=c]{70}{TECA} &
\rotatebox[origin=c]{70}{Gaussian} & \rotatebox[origin=c]{70}{shot} & \rotatebox[origin=c]{70}{impulse} & \rotatebox[origin=c]{70}{defocus} & \rotatebox[origin=c]{70}{glass} & \rotatebox[origin=c]{70}{motion} & \rotatebox[origin=c]{70}{zoom} & \rotatebox[origin=c]{70}{snow} & \rotatebox[origin=c]{70}{frost} & \rotatebox[origin=c]{70}{fog}  & \rotatebox[origin=c]{70}{brightness} & \rotatebox[origin=c]{70}{contrast} & \rotatebox[origin=c]{70}{elastic\_trans} & \rotatebox[origin=c]{70}{pixelate} & \rotatebox[origin=c]{70}{jpeg} & Mean $\downarrow$ \\ \midrule
Source &         & 95.30            & 94.58        & 95.30           & 84.86          & 91.08        & 86.84         & 77.18       & 84.38 & 79.70  & 77.26 & 44.42       & 95.58     & 85.24               & 76.92     & 66.68              & 82.35 \\
Input Adapt   &         & 83.10 & 82.62 & 83.04 & 84.12 & 90.32 & 85.08 & 77.82 & 80.34 & 74.72 & 70.90 & 44.88 & 89.34 & 81.06 & 62.18 & 59.48 & 76.60 \\
BN Adapt &       & 87.62            & 87.46        & 87.78           & 87.80          & 87.98        & 78.28         & 64.42       & 67.64 & 70.62  & 54.84 & 36.44       & 89.24     & 58.04               & 56.42     & 66.54              & 72.07 \\
\rowcolor{lightgray} BN Adapt & \checkmark & 83.24            & 82.42        & 83.48           & 86.66          & 86.10        & 75.70         & 64.28       & 66.34 & 68.00  & 54.92 & 35.98       & 82.90     & 57.36               & 54.26     & 62.42              & 69.60 \\
Tent     &       & 85.64            & 79.98          & 78.14           & 82.20          & \underline{79.22}    & 71.04          & 59.12          & 65.58          & 66.38          & 55.52          & 40.52          & 80.46          & 55.58               & 53.44          & 59.14              & 67.46          \\
                       \rowcolor{lightgray} Tent  & \checkmark & \underline{79.08}      & \underline{73.54}    & \textbf{72.92}  & \textbf{80.74} & \textbf{78.86} & \underline{69.52}    & 59.54          & 65.02          & 64.76          & 55.06          & 40.50          & 77.56          & 56.02               & 53.40          & 58.04              & 65.64          \\
EATA     &       & 84.04            & 80.16          & 80.96           & 83.52          & 82.58          & 70.64          & \underline{58.40}    & 62.00          & 65.72          & 49.80          & \textbf{34.98} & 82.20          & 53.38               & 49.94          & 59.98              & 66.55          \\
                       \rowcolor{lightgray} EATA   & \checkmark & \textbf{77.44}   & \textbf{73.26} & \underline{74.90}     & \underline{81.36}    & 81.02          & \textbf{68.96} & \textbf{58.18} & \textbf{61.06} & \underline{63.58}    & \textbf{48.74} & \underline{35.06}    & 77.74          & 52.58               & 48.82          & 56.42              & \underline{63.94}    \\
CoTTA  &       & 87.34            & 86.18          & 84.36           & 85.82          & 84.34          & 75.06          & 63.44          & 63.28          & 63.92          & 52.32          & 38.54          & \underline{73.14}    & \underline{51.06}         & \underline{45.14}    & \underline{50.38}        & 66.95          \\
                     \rowcolor{lightgray} CoTTA     & \checkmark & 81.04            & 77.68          & 76.74           & 82.24          & 79.64          & 69.56          & 59.74          & \underline{61.44}    & \textbf{61.06} & \underline{49.62}    & 38.66          & \textbf{66.16} & \textbf{49.70}      & \textbf{44.22} & \textbf{48.40}     & \textbf{63.06} \\ \bottomrule
\end{tabular}
}

\label{tab:standard}
\end{table*}

\subsubsection{Results of the Domain Generalization Benchmarks}
\cref{tab:domainbed} shows the results of the domain generalization benchmarks.
TECA improves the accuracy of T3A and shows the best results.
This result indicates the effectiveness of TECA for various types of distribution shifts.

\begin{table*}[t]
\centering
\caption{
Classification accuracy of ResNet-50 for the domain generalization benchmarks.
We experimented three times and report the mean and standard error.
}
\begin{tabular}{l|c|cccc|c}
\toprule
Method        & TECA & VLCS             & PACS             & OfficeHome       & TerraIncognita   & Mean $\uparrow$ \\
\midrule
Source                       & & 76.4 $\pm$ 0.3            & 84.5 $\pm$ 0.8            & 66.8 $\pm$ 0.1            & \underline{45.5} $\pm$ 0.4            & 68.3                      \\
Input Adapt            & & 75.8 $\pm$ 0.3            & 85.2 $\pm$ 0.7            & 66.8 $\pm$ 0.1            & \textbf{45.7} $\pm$ 1.0            & \underline{68.4}                      \\
T3A                   & & \textbf{76.8} $\pm$ 0.3            & \underline{85.4} $\pm$ 0.6            & \underline{67.8} $\pm$ 0.2            & 43.5 $\pm$ 0.4            & \underline{68.4}                      \\
\rowcolor{lightgray}T3A               & \checkmark & \underline{76.6} $\pm$ 0.3            & \textbf{86.1} $\pm$ 0.6            & \textbf{67.9} $\pm$ 0.2            & 44.6 $\pm$ 0.7            & \textbf{68.8}                      \\
\bottomrule
\end{tabular}
\label{tab:domainbed}
\end{table*}

\subsubsection{Trade-off between Error Rate and Number of Parameters}
\begin{figure}[t]
\centering
\includegraphics[width=1.0\linewidth]{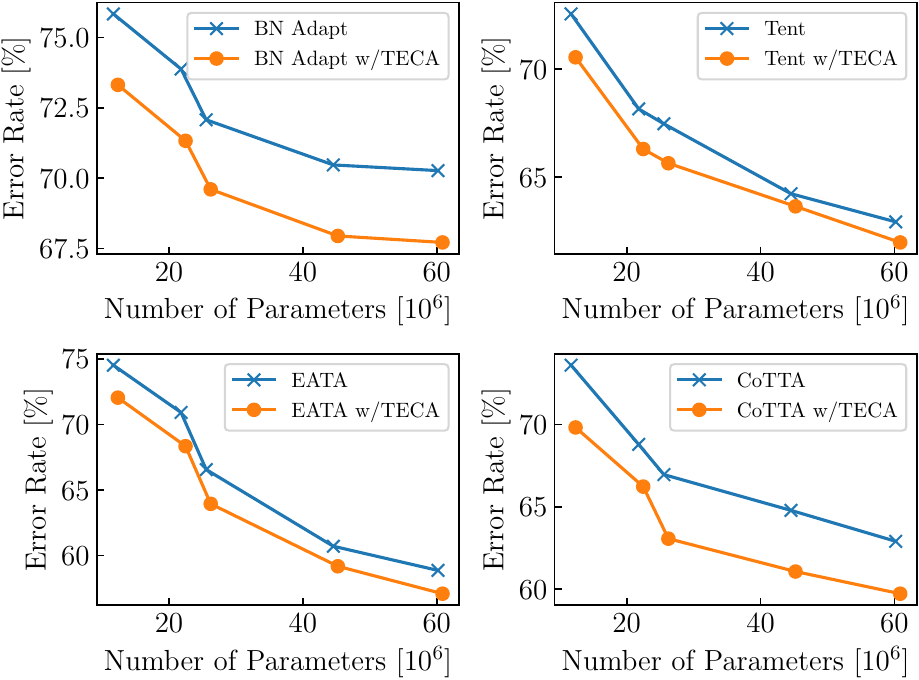}
\caption{
Trade-off between error rate and number of parameters for the standard CTTA task. 
We compared the TTA method with TECA using ResNet-18, 34, 50, 101, and 152 as classification models.
The blue and orange lines show TTA method without and with TECA.
}
\label{fig:cost_err}
\end{figure}

TECA uses an image enhancement model, which increases the number of trainable parameters.
In general, the error rate decreases as the number of trainable parameters increases.
To evaluate the parameter efficiency, we compared the TTA methods and its combination with TECA on ResNet with different numbers of parameters.
The results are shown in \cref{fig:cost_err}.
TECA improves the trade-offs of all TTA methods despite the number of parameters for URIE, the image enhancement model used in this study, being $0.67 \times 10^6$, which is sufficiently small compared with ResNet (e.g., ResNet-50 has $25.56 \times 10^6$ parameters).
The results show that using TECA is more parameter-efficient than simply increasing the number of classification model parameters.

\subsubsection{Other Architectures}
\label{sec:other}
The image enhancement model is more effective when used in conjunction with the enhancer partner because the image enhancement model is trained to minimize the loss of the enhancer partner~\cite{dyntta}.
We conducted experiments to investigate whether TECA improves the TTA methods when using a different enhancer partner than the one used when training the image enhancement model.
The results are shown in \cref{tab:other_arch}.
Note that DeiT has the same architecture as ViT~\cite{vit} and does not have a BN layer; instead we update the layer normalization layer~\cite{robustifying}.
For layer normalization layers, BN-adapt cannot be used, and TECA only works as LS.
The image enhancement model trained with ResNet-50 shows a lower error rate for similar architectures, ResNeXt-50 and Wide ResNet-50, while the error rate increases for a completely different architecture, DeiT-base.
On the other hand, the image enhancement model trained with DeiT-base slightly reduces the error rate for all architectures.
The updating of DeiT-small and DeiT-base by EATA shows catastrophic forgetting, while TECA prevents catastrophic forgetting of DeiT-small.
In most experiments, TECA reduces the error rate of the TTA methods.
The decrease in error rate for TECA is proportional to the decrease in error rate for Input Adapt.
Even when Input Adapt increases the error rate, TECA decreases the error rate of all TTA methods except Tent.
This result indicates that the image enhancement model used by TECA may be trained with any classification model and has fewer limitations for TECA.


\begin{table*}[t]
\centering
\caption{
Average classification error rate for different classification and image enhancement models for the standard CTTA task.
}
\begin{tabular}{c|c|ccccccc}
\toprule
Classifier & TECA~(Enhancer Partner) & Source & Input Adapt & BN Adapt & Tent & EATA & CoTTA \\ \midrule
 &  & 79.34 & N/A & 70.81 & 63.80 & 62.72 & 63.69 \\
 & \cellcolor{lightgray}\checkmark~(ResNet-50) & \cellcolor{lightgray}N/A & \cellcolor{lightgray}74.58 & \cellcolor{lightgray}68.67 & \cellcolor{lightgray}63.00 & \cellcolor{lightgray}\underline{60.88} & \cellcolor{lightgray}\textbf{60.43} \\
\multirow{-3}{*}{ResNeXt-50} &  \cellcolor{lightgray}\checkmark~(DeiT-base) & \cellcolor{lightgray}N/A & \cellcolor{lightgray}78.69 & \cellcolor{lightgray}69.61 & \cellcolor{lightgray}63.58 & \cellcolor{lightgray}61.23 & \cellcolor{lightgray}61.71 \\ \midrule
 &  &  79.41 & N/A & 70.22 & 62.97 & 62.54 & 63.61 \\
 & \cellcolor{lightgray}\checkmark~(ResNet-50) & \cellcolor{lightgray}N/A & \cellcolor{lightgray}74.43 &  \cellcolor{lightgray}68.15 & \cellcolor{lightgray}61.88 & \cellcolor{lightgray}\underline{60.42} & \cellcolor{lightgray}\textbf{59.94} \\
\multirow{-3}{*}{WideResNet-50} & \cellcolor{lightgray}\checkmark~(DeiT-base) & \cellcolor{lightgray}N/A & \cellcolor{lightgray}79.31 &  \cellcolor{lightgray}69.05 & \cellcolor{lightgray}62.51 & \cellcolor{lightgray}60.94 & \cellcolor{lightgray}60.86 \\ \midrule
 &  & 75.30 & N/A  & N/A & 73.50 & 72.31 & 89.22 \\
 & \cellcolor{lightgray}\checkmark~(ResNet-50) & \cellcolor{lightgray}N/A & \cellcolor{lightgray}72.96  & \cellcolor{lightgray}72.36 & \cellcolor{lightgray}70.15 & \cellcolor{lightgray}\textbf{68.77} & \cellcolor{lightgray}70.18 \\
\multirow{-3}{*}{DeiT-tiny} & \cellcolor{lightgray}\checkmark~(DeiT-base) & \cellcolor{lightgray}N/A & \cellcolor{lightgray}72.93 & \cellcolor{lightgray}72.78 & \cellcolor{lightgray}71.04 & \cellcolor{lightgray}\underline{69.17} & \cellcolor{lightgray}77.36 \\ \midrule
 &  & 62.43 & N/A  & N/A & 58.65 & 99.08 & 55.95 \\
 & \cellcolor{lightgray}\checkmark~(ResNet-50) & \cellcolor{lightgray}N/A & \cellcolor{lightgray}61.38  & \cellcolor{lightgray}60.32 & \cellcolor{lightgray}64.63 & \cellcolor{lightgray}96.90 & \cellcolor{lightgray}\underline{54.54} \\
\multirow{-3}{*}{DeiT-small} & \cellcolor{lightgray}\checkmark~(DeiT-base) & \cellcolor{lightgray}N/A & \cellcolor{lightgray}60.05  & \cellcolor{lightgray}59.80 & \cellcolor{lightgray}57.47 & \cellcolor{lightgray}54.66 & \cellcolor{lightgray}\textbf{53.49} \\ \midrule
 &  &  55.04 & N/A & N/A & 51.25 & 99.72 & 50.89 \\
 & \cellcolor{lightgray}\checkmark~(ResNet-50) & \cellcolor{lightgray}N/A & \cellcolor{lightgray}56.98 & \cellcolor{lightgray}54.30 & \cellcolor{lightgray}52.16 & \cellcolor{lightgray}99.65 & \cellcolor{lightgray}\underline{49.81} \\
\multirow{-3}{*}{DeiT-base} & \cellcolor{lightgray}\checkmark~(DeiT-base) & \cellcolor{lightgray}N/A & \cellcolor{lightgray}54.15 & \cellcolor{lightgray}52.95 & \cellcolor{lightgray}50.58 & \cellcolor{lightgray}99.71 & \cellcolor{lightgray}\textbf{49.48} \\ \bottomrule
\end{tabular}
\label{tab:other_arch}
\end{table*}

\subsection{More Discussion}
In this section, we conducted more detailed experiments on the standard CTTA task.

\subsubsection{Range for Updating Parameters}
\label{sec:param_range}
Although TECA updates both the classification and image enhancement models, we evaluated the error rate when only the respective models are updated.
The results are shown in \cref{tab:update_params}.
We found that updating only the image enhancement model with BN Adapt results in an increased error rate.
Since the image enhancement model improves classification accuracy by transforming corrupted images into ones that are closer to the source domain used for training, the results suggest that source domain knowledge, especially BN statistics, is highly important.
For all other TTA methods except EATA, the error rate is increased by updating only the image enhancement model despite freezing the BN statistics.
This result indicates that updating the image enhancement model from uncertain predictions with high entropy has a significantly negative impact.
Updating only the classification model with TECA reduces the error rate for all TTA methods.
Updating both models has the best results, even though updating only the image enhancement model increased the error rate.
The average entropy when updating only the image enhancement model for Input Adapt, BN Adapt, Tent, EATA, and CoTTA was 3.31, 3.03, 2.88, 3.03, and 3.03, respectively.
Although updating only the image enhancement model with TECA decreases accuracy, it also reduces average entropy.
Updating the model with high-entropy predictions causes error accumulation~\cite{eata}, while our method improves the stability of TTA methods by lowering the entropy of the predictions.


\begin{table}[t]
\centering
\caption{
Average classification error rate of ResNet-50 when changing the parameter to be updated.
The numbers in parentheses indicate the difference from when the model is not updated.
}
\resizebox{\columnwidth}{!}{
\begin{tabular}{ccccc}
\toprule
Update & BN Adapt & Tent & EATA & CoTTA \\ \midrule
Enhancer $\phi$ & 77.85\color{red}(1.25) & 77.60\color{red}(1.00) & 76.56\color{green}(-0.04) & 77.41\color{red}(0.81) \\
Classifier $\theta$ & 69.66\color{green}(-6.94) & 65.85\color{green}(-10.75) & 64.13\color{green}(-12.47) & 63.52\color{green}(-13.08) \\
Both $\Theta$ & 69.60\color{green}(-7.00) & 65.64\color{green}(-10.96) & 63.94\color{green}(-12.66) & 63.06\color{green}(-13.54) \\ \bottomrule
\end{tabular}
}
\label{tab:update_params}
\end{table}


\subsubsection{Ablation Studies}
We evaluated the effectiveness of each of the three modules that further increase accuracy beyond simply combining and updating the classification and image enhancement models.
The results are shown in \cref{tab:ablation_study}.
The improvement in error rate due to LS is significant, and in EATA, in particular, the model collapsed without LS.
SPUS shows a slight improvement for Tent and EATA, while CoTTA decreases the error rate.
Unlike other TTA methods, CoTTA uses all model parameters for training, so the impact of parameter update speed is significant.
As described in \cref{sec:param_range}, BN Adapt is not suitable for training the image enhancement model, and removing BN Adapt from the TTA method reduces the error rate.


\begin{table}[t]
\centering
\caption{
Average classification error rate of ResNet-50 when each of the three modules is excluded.
The numbers in parentheses indicate the differences from the TECA with all modules included.
BN Adapt requires updating the BN statistics and does not update the trainable parameters, so SPUS and FBNS are not excluded.
}
\resizebox{\columnwidth}{!}{
\begin{tabular}{ccccc}
\toprule
Excluded & BN Adapt & Tent & EATA & CoTTA \\ \midrule
 LS & 71.39\color{red}(1.79) & 66.14\color{red}(0.50) & 99.52\color{red}(35.58) & 63.94\color{red}(0.88) \\
 SPUS & N/A & 65.68\color{red}(0.04) & 63.95(0.00) & 63.67\color{red}(0.61) \\
 FBNS & N/A & 66.22\color{red}(0.58) & 64.17\color{red}(0.23) & 63.87\color{red}(0.81) \\ \bottomrule
\end{tabular}
}

\label{tab:ablation_study}
\end{table}




\section{Conclusion}
\label{sec:conc}
In this paper, we propose a new problem setting that combines and updates the classification model and the image enhancement model during testing, and a novel method, TECA, that further increases accuracy of existing TTA methods by updating both models from low-uncertainty predictions.
To further increase the accuracy of TECA, we introduce two modules: SPUS and FBNS.
Although TECA does not have hyperparameters and the overhead of trainable parameters is very small, its training efficiency is higher than simply increasing the number of the classification model parameters.
We evaluated TECA in the continual TTA task and the domain generalization benchmarks.
We find that TECA further increases the accuracy of state-of-the-art TTA methods in all experimental settings.
Furthermore, more detailed experiments show the parameter efficiency and validity of TECA and that it works with a variety of classification models.



\bibliographystyle{IEEEtran}
\bibliography{egbib}

\end{document}


\title{Supplementary Materials for Test-time Adaptation Meets Image Enhancement: Improving Accuracy via Uncertainty-aware Logit Switching}


\maketitle

\setcounter{table}{6}
\setcounter{figure}{4}
\setcounter{equation}{2}
\setcounter{section}{5}

\begin{abstract}
The supplementary materials for ``Test-time Adaptation Meets Image Enhancement: Improving Accuracy via Uncertainty-aware Logit Switching.''
The additional experimental results are provided.

\end{abstract}

\section{CTTA Diverse and Gradual Tasks}
In the diverse task, we evaluated the average of 10 trials in different evaluation orders of the corrupted test set.
In the gradual task, the evaluation order of the corrupted test set is the same as in the standard task, and the corruption severity level is varied in steps of $1{\xrightarrow{}}2{\xrightarrow{}}3{\xrightarrow{}}4{\xrightarrow{}}5{\xrightarrow{}}4{\xrightarrow{}}3{\xrightarrow{}}2{\xrightarrow{}}1$.

Table~\ref{tab:diverse} shows the results of the diverse task .
As with the standard task, TECA consistently improves the average error rate of the TTA methods.
Due to the irregularity of the corruption order in the diverse task, CoTTA, which is robust against large distribution shifts, performs well.
Furthermore, the combination of CoTTA and TECA reduces the average error rate by $2.01$ points over CoTTA alone.

\begin{table}[tb]
\centering
\caption{
Average classification error rate of ResNet-50 for the diverse ImageNet CTTA task. 
}
\resizebox{\columnwidth}{!}{
\begin{tabular}{c|cccccc}
\toprule
TECA & Source & Input Adapt & BN Adapt & Tent  & EATA  & CoTTA \\ \midrule
            & 82.35  & 76.60 & 72.07    & 66.52 & 66.61 & \underline{63.20} \\
\rowcolor{lightgray} \checkmark             &        &       & 69.60    & 65.13 & 64.20 & \textbf{61.29} \\ \midrule  

\end{tabular}
}

\label{tab:diverse}
\end{table}
\begin{table}[tb]
\centering
\caption{
Average classification error rate of ResNet-50 for the gradual ImageNet CTTA task. 
}

\resizebox{\columnwidth}{!}{
\begin{tabular}{c|cccccc}
\toprule
TECA & Source & Input Adapt  & BN Adapt & Tent  & EATA  & CoTTA \\ \midrule
            & 59.61 & 54.86 & 51.41 & 52.09 & 46.70 & \underline{40.66} \\
\rowcolor{lightgray} \checkmark             &        &       & 49.73 & 57.20 & 45.38 & \textbf{39.23} \\ \bottomrule  
\end{tabular}
}

\label{tab:gradual}
\end{table}

Table~\ref{tab:gradual} shows the results of the gradual task.
With the exception of Tent, TECA improves the average error rate of the TTA methods.
Tent shows a higher error rate than BN Adapt, which indicates that training the BN parameters by Tent is detrimental in long-term adaptation tasks.
The error rate of TECA is higher than that of the original Tent because TECA increases the number of BN parameters to be trained and the impact of error accumulation is greater.
As with the diverse tasks, CoTTA performs better than the other TTA methods, with the best results obtained by combining it with TECA.

From these CTTA task results, we found that a combination of TECA and TTA methods improves performance more in various situations.

\section{Results of the PTTA Task}
In the PTTA task, we used Dirichlet distributions to generate streaming data that are correlated between test samples and simulated real-world applications.
For the PTTA task, we use \textit{RoTTA}~\cite{rotta}, a TTA method that allows stable adaptation by using a memory bank under correlated non-i.i.d. data streams.

We show the results of the PTTA task in Table~\ref{tab:ptta}.
TECA increases the accuracy even for RoTTA, the state-of-the-art method in realistic tasks.
Combining RoTTA with TECA shows the best results, reducing the error rate by $0.99$ points over RoTTA alone.
For late-time corruptions (pixelate and jpeg), TECA increases the RoTTA error rate.
Incorporating a method such as CoTTA, which is more robust against long-term, drastically changing distribution shifts, may solve this problem.

\begin{table*}[t]
\centering
\caption{
Classification error rate of ResNet-50 for the ImageNet PTTA task. 
}
\resizebox{\textwidth}{!}{
\begin{tabular}{l|c|lllllllllllllll|c} 
\toprule
\multicolumn{2}{l|}{Time} & \multicolumn{15}{l|}{$t\xrightarrow{\hspace*{15.0cm}}$}& \\ \midrule
        Method & \rotatebox[origin=c]{70}{TECA} & \rotatebox[origin=c]{70}{motion} & \rotatebox[origin=c]{70}{snow} & \rotatebox[origin=c]{70}{fog} & \rotatebox[origin=c]{70}{shot} & \rotatebox[origin=c]{70}{defocus} & \rotatebox[origin=c]{70}{contrast} & \rotatebox[origin=c]{70}{zoom} & \rotatebox[origin=c]{70}{brightness} & \rotatebox[origin=c]{70}{frost} & \rotatebox[origin=c]{70}{elastic} & \rotatebox[origin=c]{70}{glass} & \rotatebox[origin=c]{70}{Gaussian} & \rotatebox[origin=c]{70}{pixelate} & \rotatebox[origin=c]{70}{jpeg} & \rotatebox[origin=c]{70}{impulse}
        & Mean $\downarrow$ \\ \midrule
Source &  & 86.82 & 84.37 & 77.25 & 94.56 & \underline{84.83} & 95.57 & 77.15 & 44.34 & 79.74 & 85.21 & 91.09 & 95.30 & 76.91 & 66.65 & 95.30 & 82.34 \\
Input Adapt &  & 85.05 & 80.32 & 70.89 & \textbf{82.56} & \textbf{84.13} & 89.31 & 77.77 & 44.84 & 74.76 & 81.02 & 90.33 & \textbf{83.05} & \underline{62.16} & \textbf{59.43} & \textbf{83.01} & 76.57 \\
RoTTA &  & \underline{79.53} & \underline{72.27} & \underline{57.30} & 88.80 & 87.74 & \underline{81.88} & \underline{66.77} & \underline{37.88} & \underline{67.88} & \underline{59.75} & \textbf{83.85} & 88.60 & \textbf{60.77} & \underline{65.15} & 88.92 & \underline{72.47} \\
 \rowcolor{lightgray} RoTTA & \checkmark & \textbf{75.92} & \textbf{68.02} & \textbf{55.06} & \underline{83.25} & 85.85 & \textbf{77.67} & \textbf{64.71} & \textbf{37.80} & \textbf{66.19} & \textbf{59.57} & \underline{84.01} & \underline{84.99} & 64.97 & 75.70 & \underline{88.50} & \textbf{71.48} \\ \bottomrule
\end{tabular}
}

\label{tab:ptta}
\end{table*}

\section{Detailed Results of the Domain Generalization Benchmarks}
Tables~\ref{tab:vlcs}--\ref{tab:terraincognita} show the results for each domain generalization dataset.

\begin{table*}[t]
\centering
\caption{
Classification accuracy of ResNet-50 on the VLCS dataset.
}
\begin{tabular}{l|c|cccc|c}
\toprule
Method   & TECA & C         & L           & S           & V           & Mean $\uparrow$ \\
\midrule
Source                  & & 97.3 $\pm$ 0.3       & 64.0 $\pm$ 0.6       & 71.1 $\pm$ 0.8       & 73.2 $\pm$ 0.5       & 76.4                 \\
Input Adapt       & & 97.4 $\pm$ 0.7       & 63.8 $\pm$ 0.2       & 69.0 $\pm$ 1.0       & 73.0 $\pm$ 0.7       & 75.8                 \\
T3A              & & 98.6 $\pm$ 0.1       & 64.7 $\pm$ 0.5       & 71.4 $\pm$ 1.0       & 72.7 $\pm$ 0.7       & 76.8                 \\
\rowcolor{lightgray}T3A          & \checkmark & 98.2 $\pm$ 0.3       & 63.8 $\pm$ 0.4       & 71.7 $\pm$ 1.7       & 72.8 $\pm$ 0.9       & 76.6                 \\
\bottomrule

\end{tabular}
\label{tab:vlcs}
\centering 
\caption{ 
Classification accuracy of ResNet-50 on the PACS dataset. 
} 
\begin{tabular}{l|c|cccc|c}
\toprule
Method   & TECA & A           & C           & P           & S           & Mean $\uparrow$\\
\midrule
Source                  & & 82.2 $\pm$ 0.6       & 80.7 $\pm$ 1.5       & 96.6 $\pm$ 0.5       & 78.4 $\pm$ 1.4       & 84.5                 \\
Input Adapt       & & 84.0 $\pm$ 0.3       & 80.1 $\pm$ 1.2       & 96.4 $\pm$ 0.3       & 80.5 $\pm$ 1.1       & 85.2                 \\
T3A              & & 83.2 $\pm$ 0.6       & 82.0 $\pm$ 1.2       & 96.9 $\pm$ 0.3       & 79.5 $\pm$ 1.4       & 85.4                 \\
\rowcolor{lightgray}T3A          & \checkmark & 84.8 $\pm$ 0.4       & 81.9 $\pm$ 1.0       & 96.9 $\pm$ 0.4       & 80.8 $\pm$ 1.3       & 86.1                 \\
\bottomrule

\end{tabular}
\label{tab:pacs}
\centering 
\caption{
Classification accuracy of ResNet-50 on the OfficeHome dataset.
} 
\begin{tabular}{l|c|cccc|c}
\toprule
Method   & TECA & A           & C           & P           & R           & Mean $\uparrow$\\
\midrule
Source                  & & 60.6 $\pm$ 0.8       & 54.2 $\pm$ 0.4       & 75.5 $\pm$ 0.4       & 76.9 $\pm$ 0.3       & 66.8                 \\
Input Adapt       & & 60.6 $\pm$ 0.5       & 54.6 $\pm$ 0.2       & 75.2 $\pm$ 0.4       & 76.6 $\pm$ 0.1       & 66.8                 \\
T3A              & & 60.9 $\pm$ 0.4       & 55.8 $\pm$ 0.5       & 77.1 $\pm$ 0.6       & 77.6 $\pm$ 0.3       & 67.8                 \\
\rowcolor{lightgray}T3A          & \checkmark & 61.4 $\pm$ 0.4       & 56.0 $\pm$ 0.4       & 77.1 $\pm$ 0.5       & 77.3 $\pm$ 0.2       & 67.9                 \\
\bottomrule

\end{tabular}
\label{tab:officehome}
\centering 
\caption{ 
Classification accuracy of ResNet-50 on the TerraIncognita dataset.
} 
\begin{tabular}{l|c|cccc|c}
\toprule
Method   & TECA & L100      & L38         & L43         & L46         & Mean $\uparrow$\\
\midrule
Source                  & & 46.4 $\pm$ 2.4       & 44.6 $\pm$ 1.8       & 51.1 $\pm$ 0.9       & 40.0 $\pm$ 0.8       & 45.5                 \\
Input Adapt       & & 48.2 $\pm$ 2.1       & 43.6 $\pm$ 1.2       & 52.4 $\pm$ 0.1       & 38.6 $\pm$ 1.3       & 45.7                 \\
T3A              & & 43.2 $\pm$ 2.0       & 46.0 $\pm$ 0.8       & 46.6 $\pm$ 1.0       & 38.2 $\pm$ 0.5       & 43.5                 \\
\rowcolor{lightgray}T3A          & \checkmark & 47.6 $\pm$ 1.7       & 45.5 $\pm$ 1.4       & 47.6 $\pm$ 1.0       & 37.6 $\pm$ 1.0       & 44.6                 \\
\bottomrule
\end{tabular}
\label{tab:terraincognita}
\end{table*}

\section{Computational Time}
\begin{figure}[t]
\centering
\includegraphics[width=1.0\linewidth]{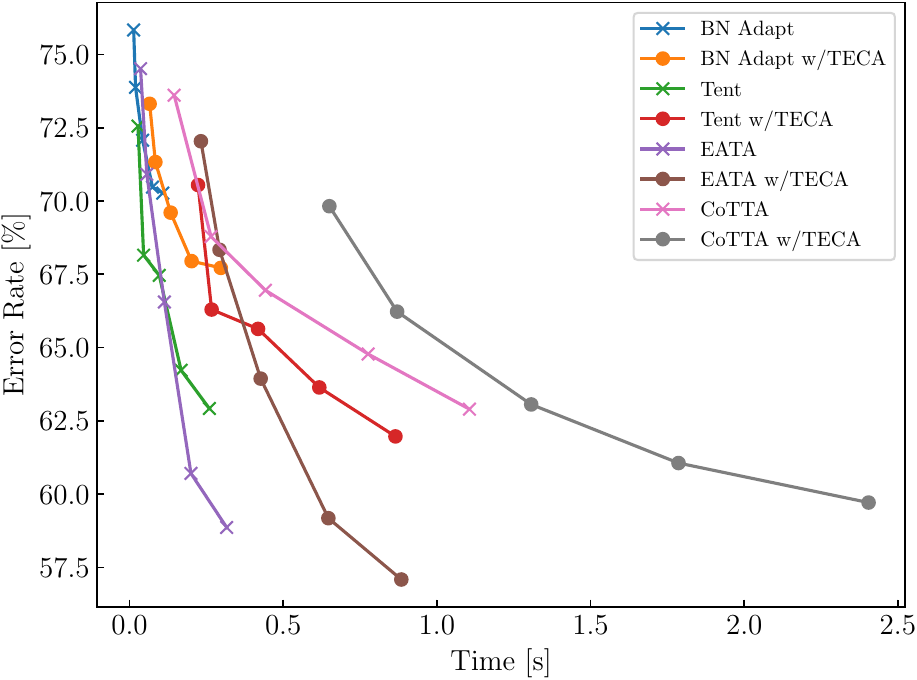}
\caption{
Trade-off between error rate and computation time for the standard ImageNet CTTA task. 
We compared the TTA method with TECA using ResNet-18, 34, 50, 101, and 152 as classification models.
}
\label{fig:time_err}
\end{figure}

TECA requires more computation time than the original TTA method because it performs inference of the image enhancement model and of the classification model for the enhanced images.
We measured the computation time and error rate for the original TTA methods and their combination with TECA, which are shown in Figure~\ref{fig:time_err}.
We experimented with a single NVIDIA-A100 GPU in the same setup as Figure~4 in main paper.
Combining TECA decreases the error rate for all methods, but increases the computation time.
However, the combination of Tent and TECA improves the tradeoff between error rate and computation time over CoTTA, the state-of-the-art TTA method.

\section{Effects of the Image Enhancement Model and LS on Confidence Scores}
\begin{figure}[t]
\centering
\includegraphics[width=1.0\linewidth]{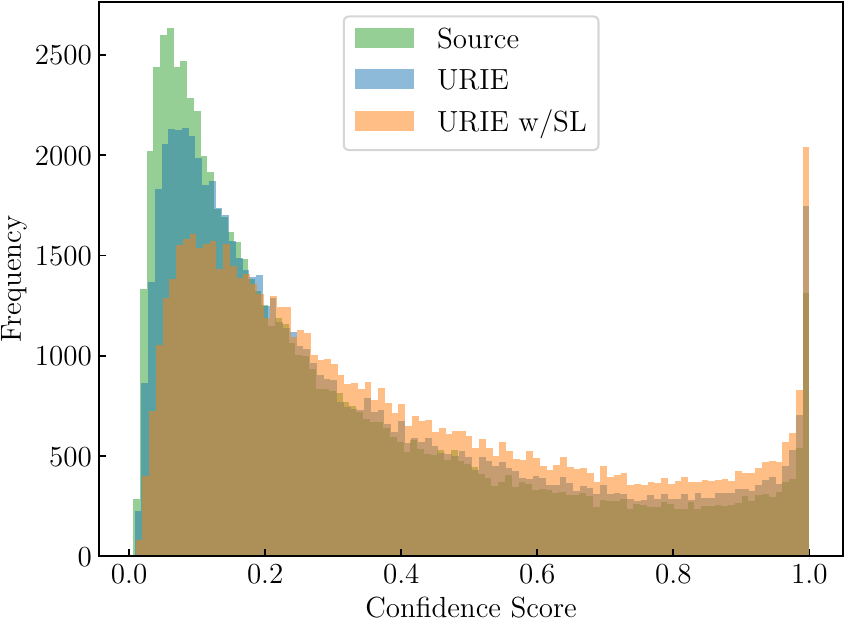}
\caption{
Confidence score histograms for each approach: Source, Input Adapt by URIE, and Input Adapt by URIE combined with LS.
We used the ResNet-50 classification model with in the standard ImageNet CTTA task. 
}
\label{fig:conf_hist}
\end{figure}

We show a histogram of confidence scores in Figure~\ref{fig:conf_hist} to demonstrate the effects of the image enhancement model and LS.
The image enhancement model reduces the number of confidence scores around 0, and LS further reduces this.
The image enhancement model and LS contribute to obtaining pseudo-labels that are closer to hard labels by increasing the number of confidence scores around 1.0.

\section{Visualization of Enhanced Images}
\begin{figure}[t]
\centering
\includegraphics[width=1.0\linewidth]{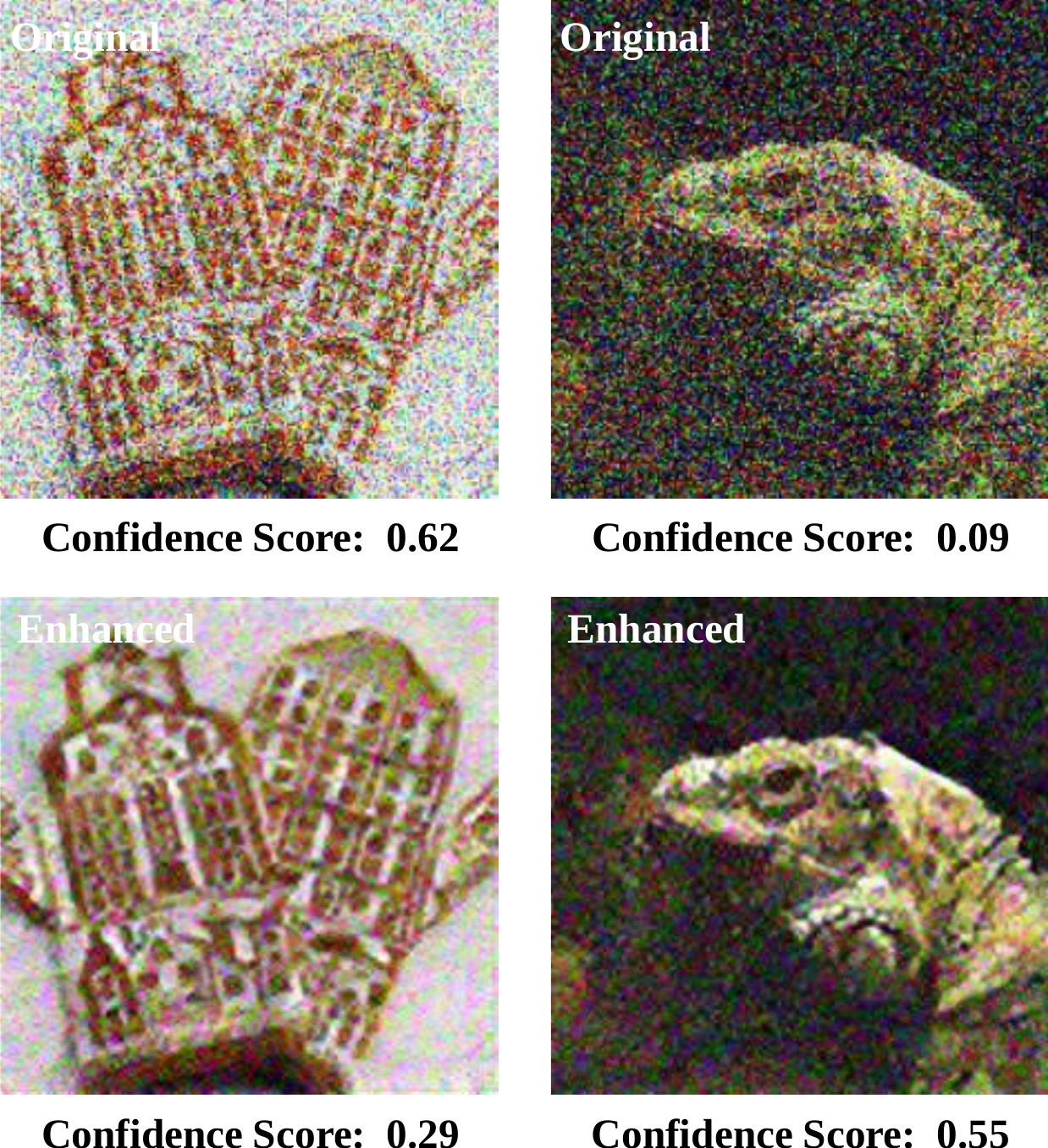}
\caption{
Gaussian noise at severity level 5 images and URIE enhanced images and their confidence scores in the ImageNet-C dataset.
The top and bottom rows are the original and enhanced images, respectively.
}
\label{fig:urie_output_all_001}
\end{figure}

\begin{figure}[t]
\centering
\includegraphics[width=1.0\linewidth]{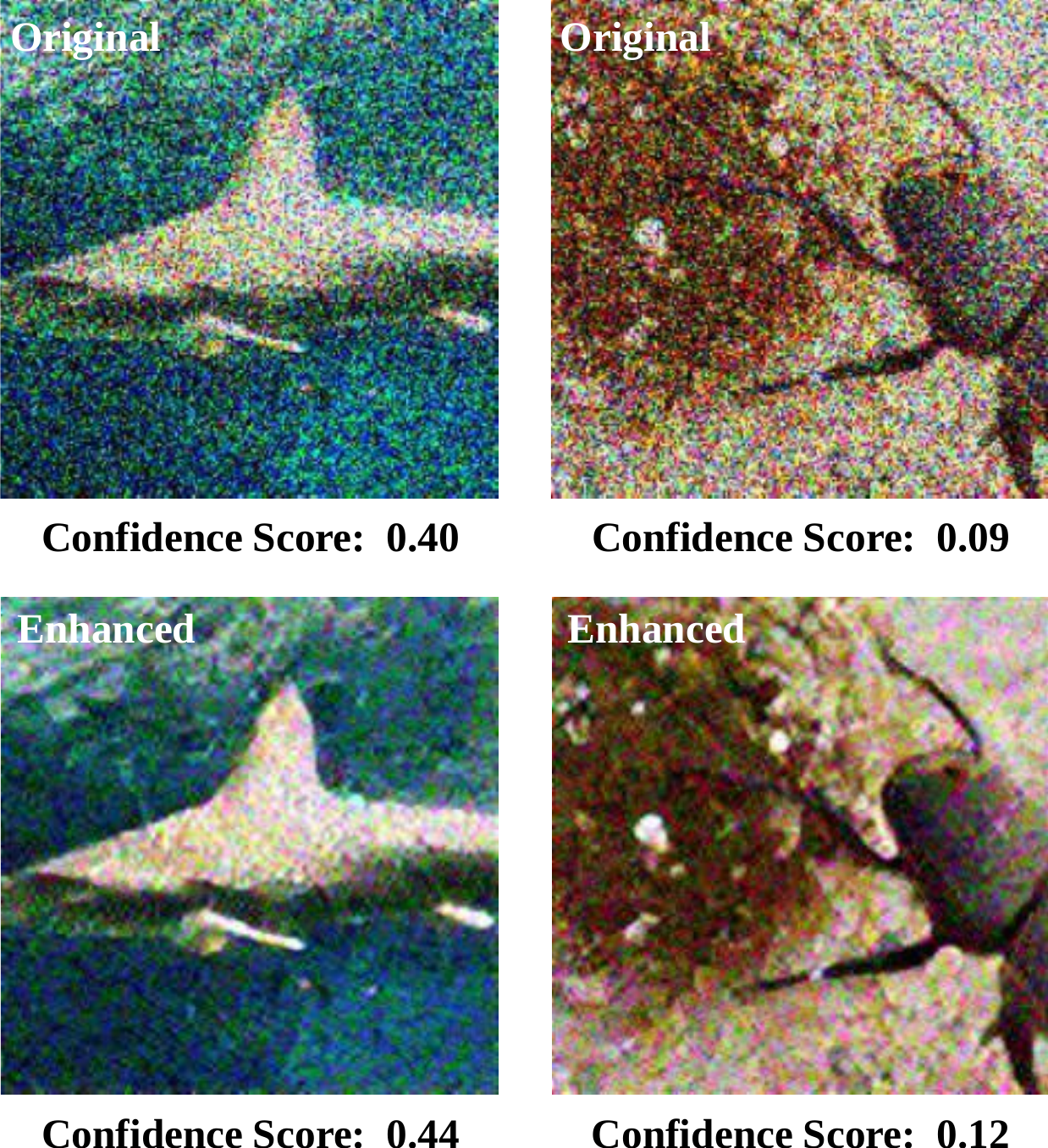}
\caption{
Shot noise at severity level 5 images and URIE enhanced images and their confidence scores in the ImageNet-C dataset.
The top and bottom rows are the original and enhanced images, respectively.
}
\label{fig:urie_output_all_002}
\end{figure}

\begin{figure}[t]
\centering
\includegraphics[width=1.0\linewidth]{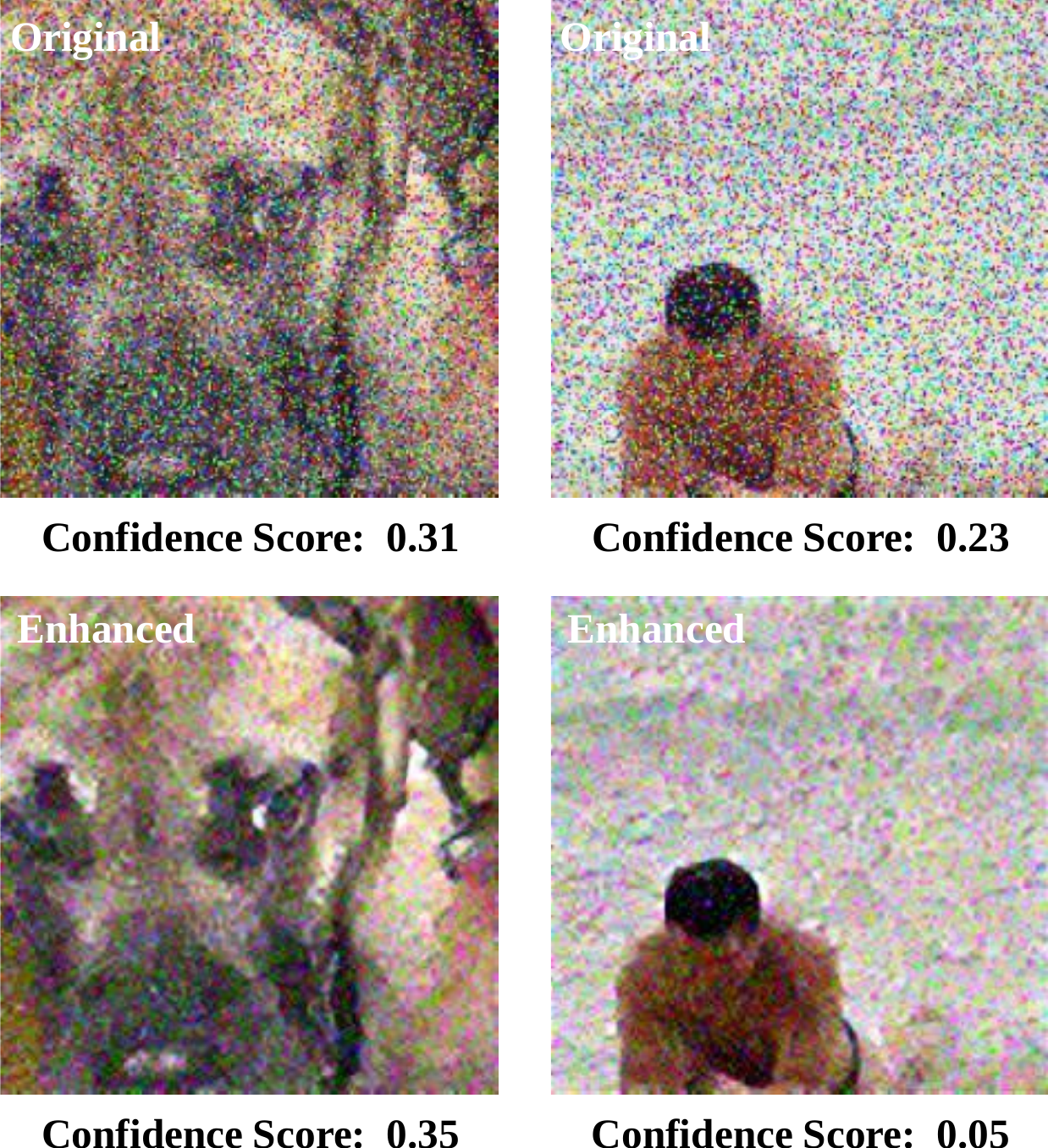}
\caption{
Impulse noise at severity level 5 images and URIE enhanced images and their confidence scores in the ImageNet-C dataset.
The top and bottom rows are the original and enhanced images, respectively.
}
\label{fig:urie_output_all_003}
\end{figure}

\begin{figure}[t]
\centering
\includegraphics[width=1.0\linewidth]{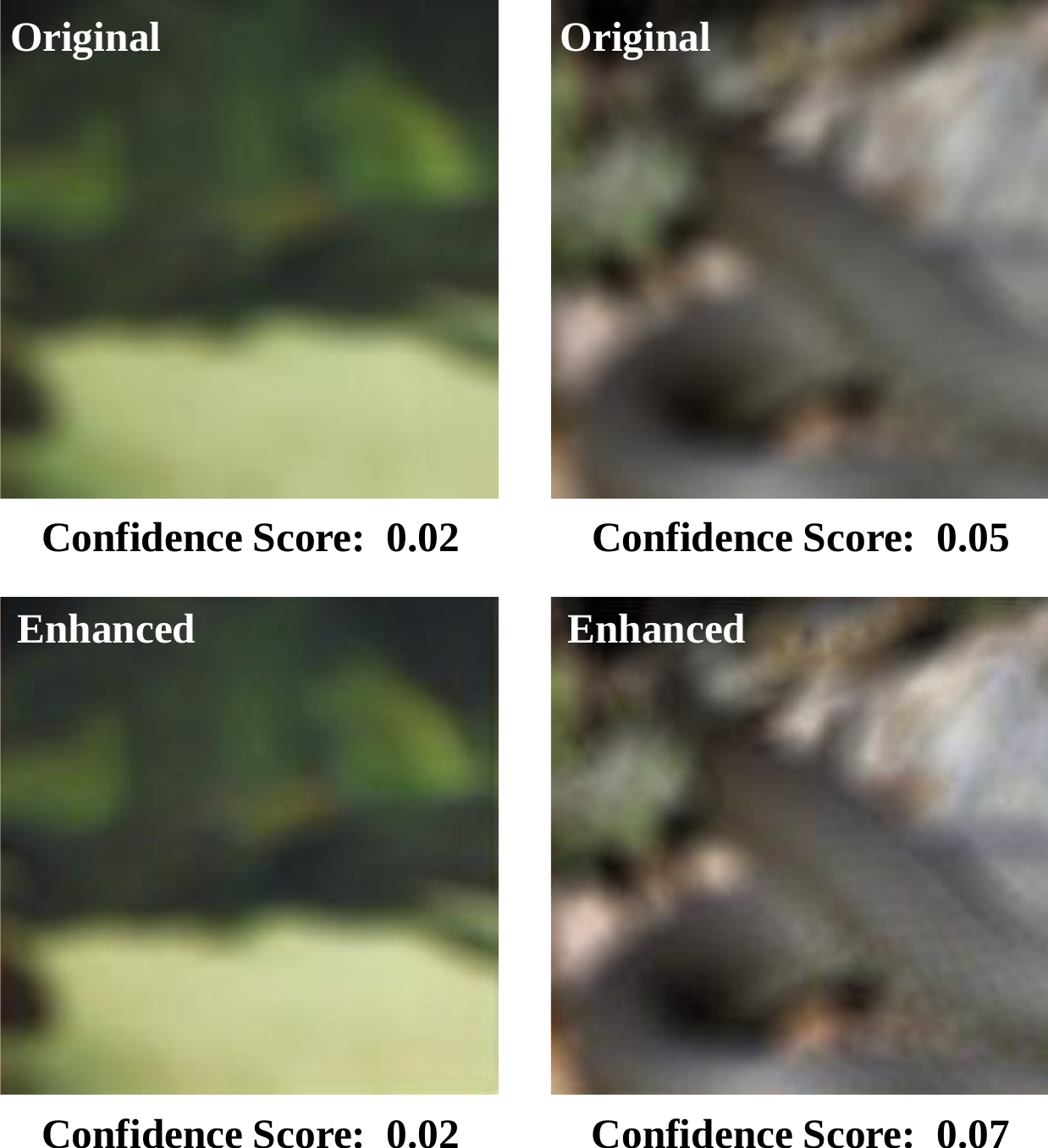}
\caption{
Defocus blur at severity level 5 images and URIE enhanced images and their confidence scores in the ImageNet-C dataset.
The top and bottom rows are the original and enhanced images, respectively.
}
\label{fig:urie_output_all_004}
\end{figure}

\begin{figure}[t]
\centering
\includegraphics[width=1.0\linewidth]{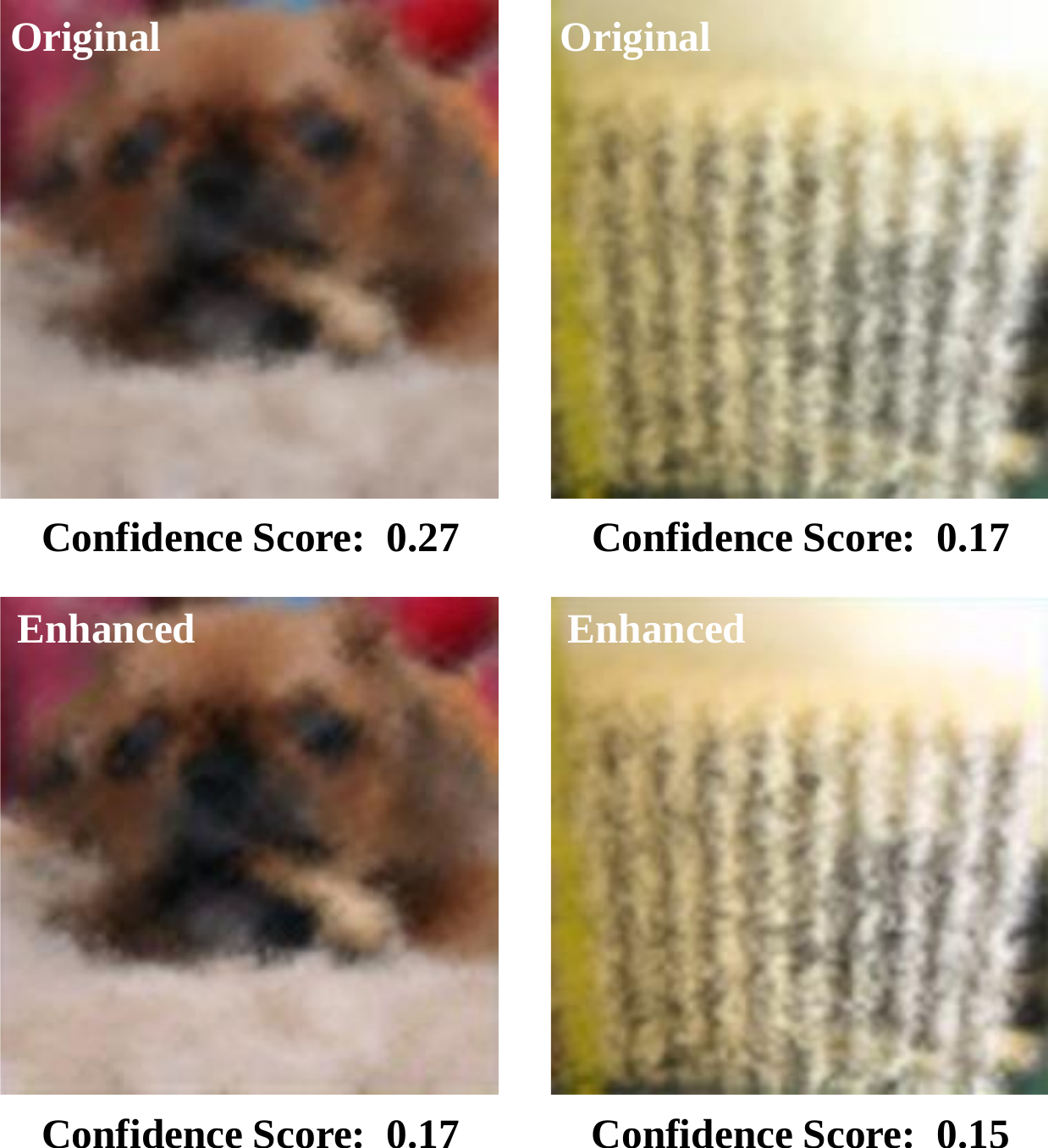}
\caption{
Glass blur at severity level 5 images and URIE enhanced images and their confidence scores in the ImageNet-C dataset.
The top and bottom rows are the original and enhanced images, respectively.
}
\label{fig:urie_output_all_005}
\end{figure}

\begin{figure}[t]
\centering
\includegraphics[width=1.0\linewidth]{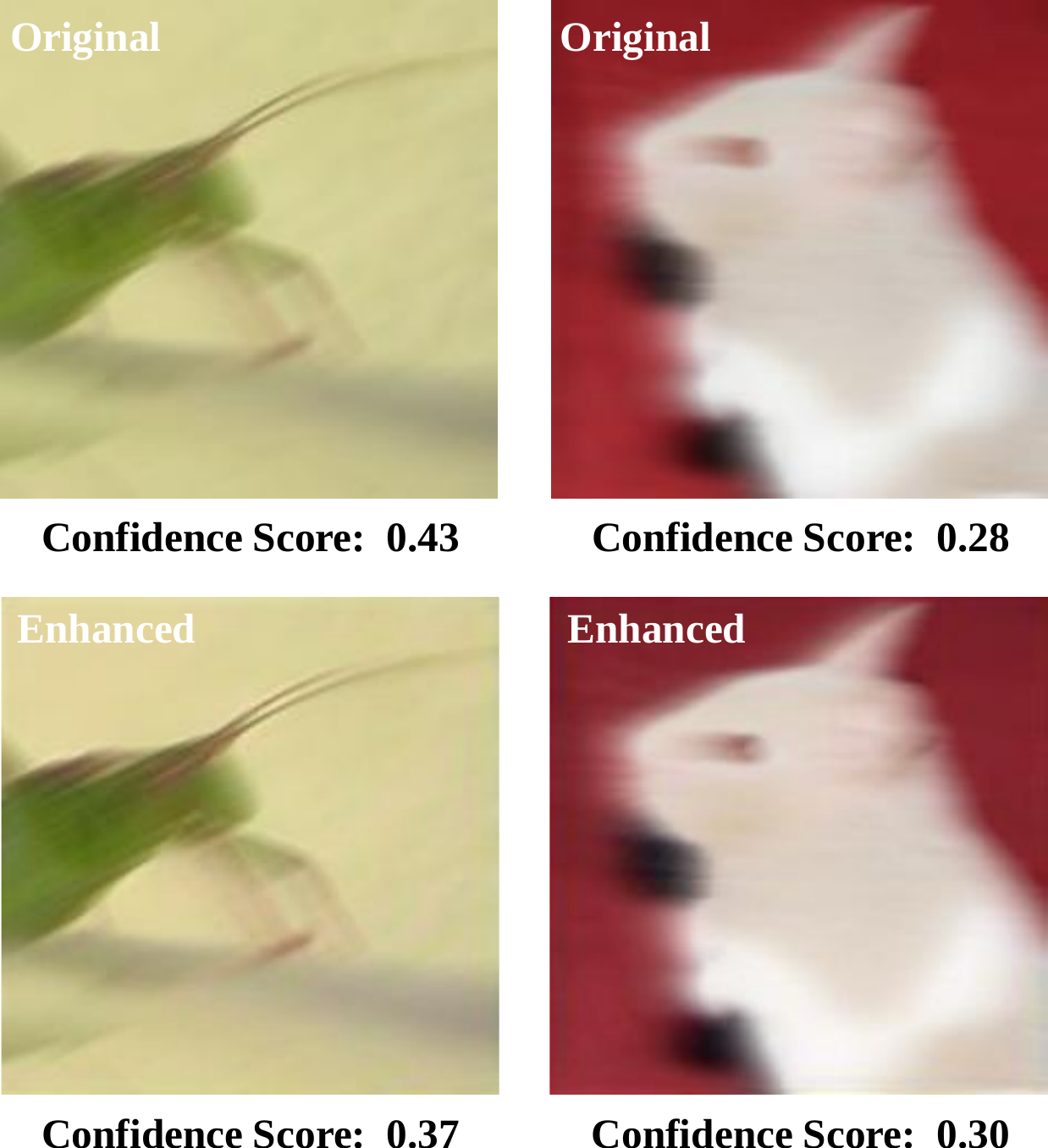}
\caption{
Motion blur at severity level 5 images and URIE enhanced images and their confidence scores in the ImageNet-C dataset.
The top and bottom rows are the original and enhanced images, respectively.
}
\label{fig:urie_output_all_006}
\end{figure}

\begin{figure}[t]
\centering
\includegraphics[width=1.0\linewidth]{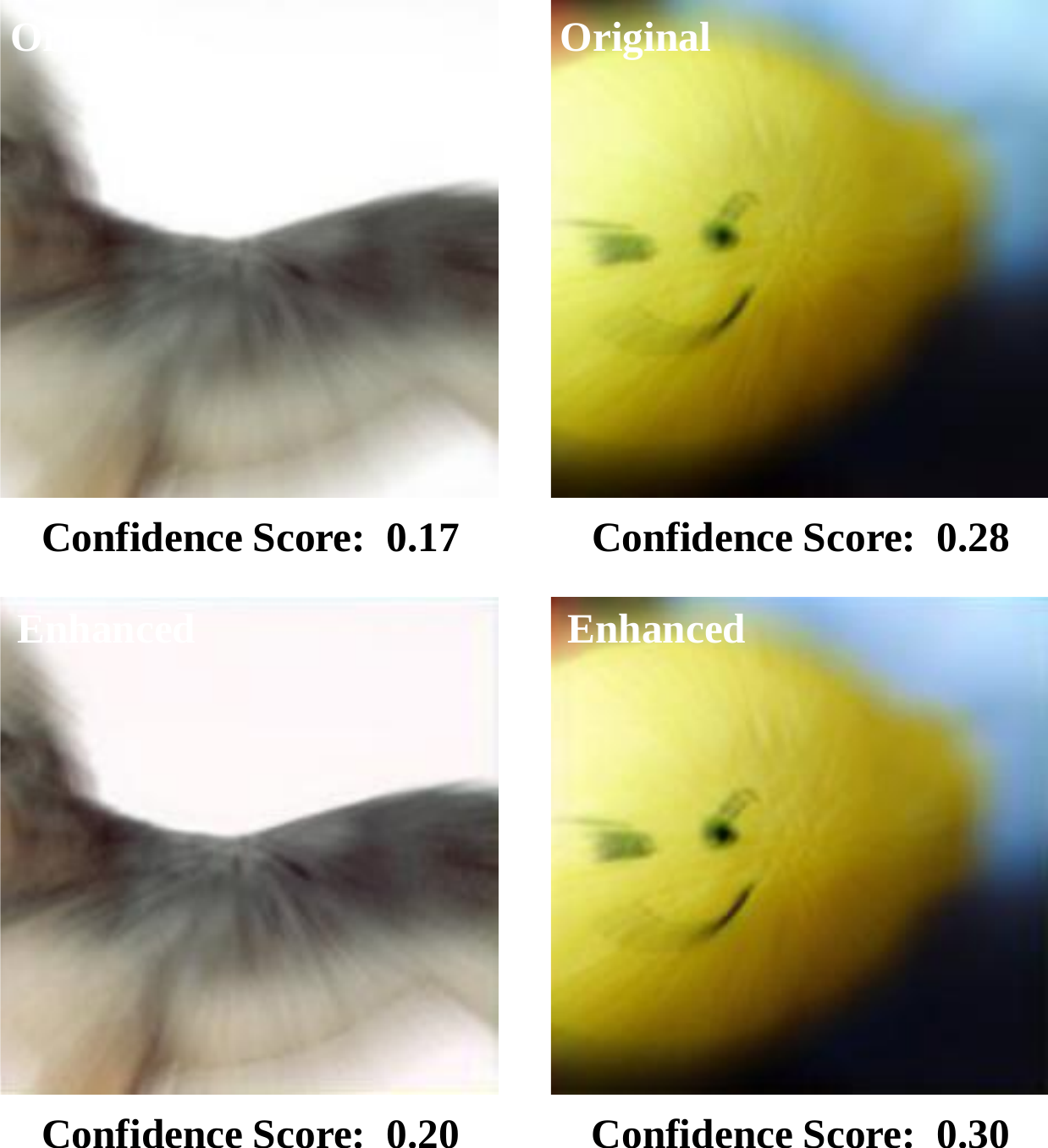}
\caption{
Zoom blur at severity level 5 images and URIE enhanced images and their confidence scores in the ImageNet-C dataset.
The top and bottom rows are the original and enhanced images, respectively.
}
\label{fig:urie_output_all_007}
\end{figure}

\begin{figure}[t]
\centering
\includegraphics[width=1.0\linewidth]{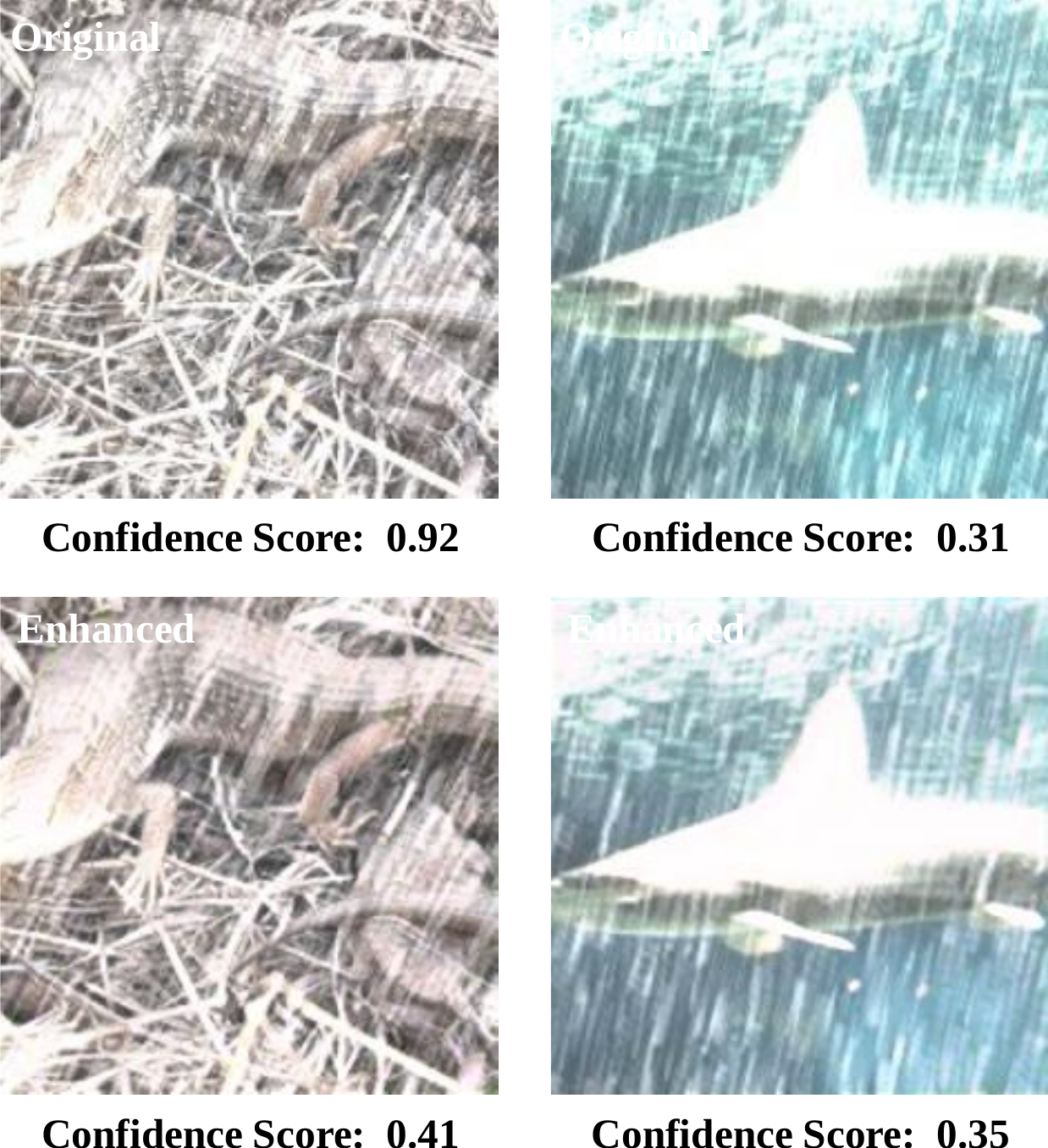}
\caption{
Snow at severity level 5 images and URIE enhanced images and their confidence scores in the ImageNet-C dataset.
The top and bottom rows are the original and enhanced images, respectively.
}
\label{fig:urie_output_all_008}
\end{figure}

\begin{figure}[t]
\centering
\includegraphics[width=1.0\linewidth]{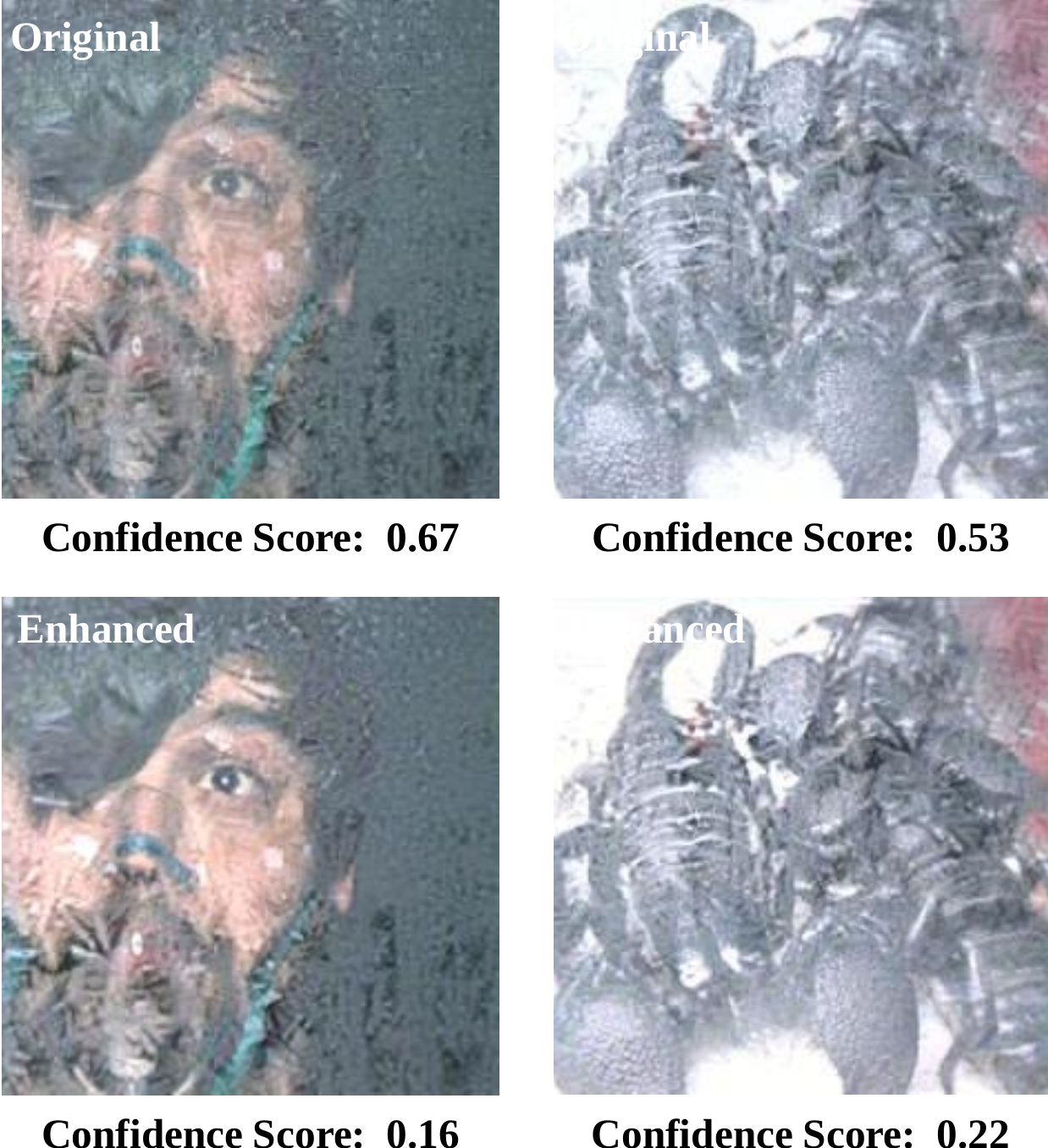}
\caption{
Frost at severity level 5 images and URIE enhanced images and their confidence scores in the ImageNet-C dataset.
The top and bottom rows are the original and enhanced images, respectively.
}
\label{fig:urie_output_all_009}
\end{figure}

\begin{figure}[t]
\centering
\includegraphics[width=1.0\linewidth]{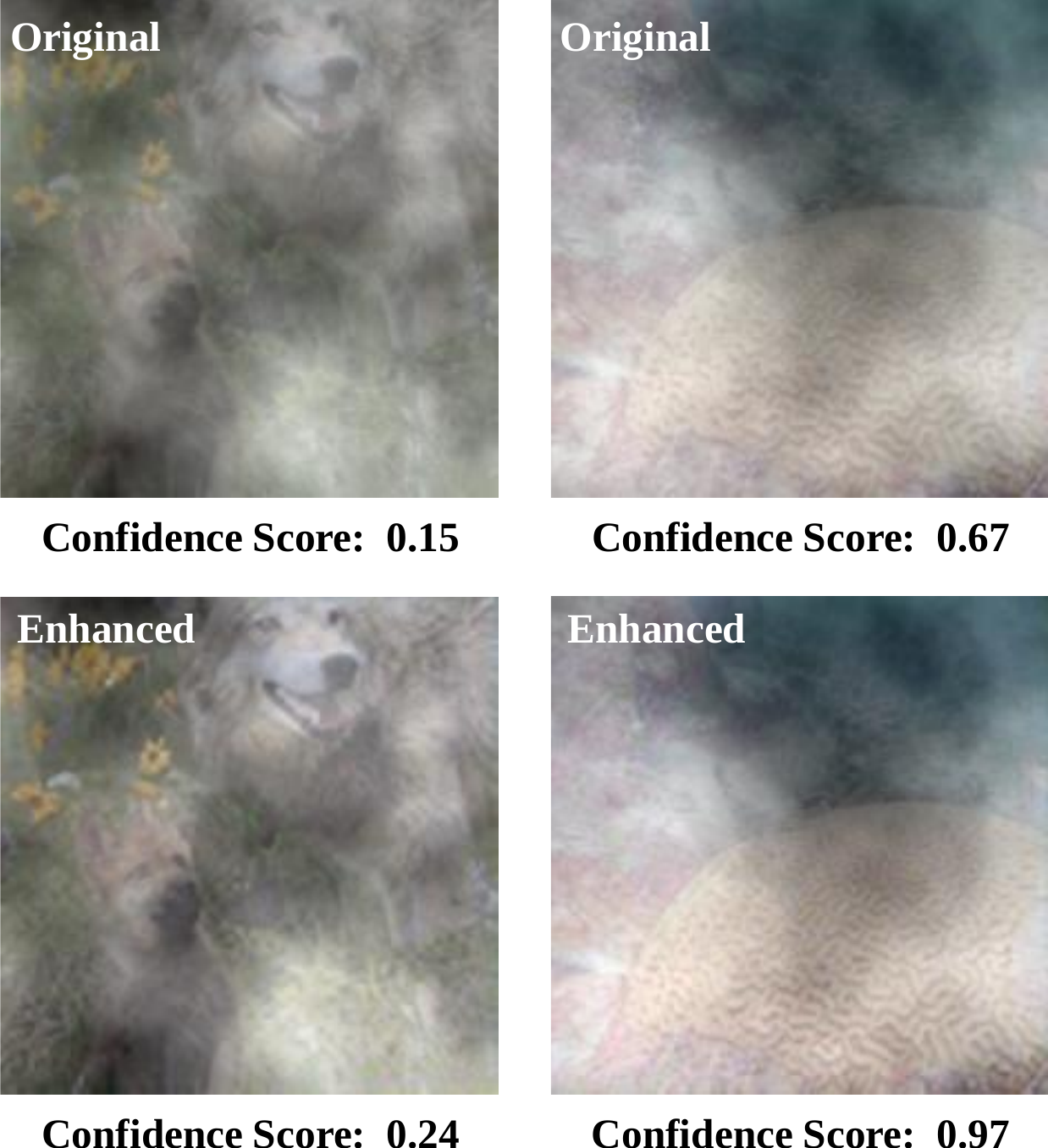}
\caption{
Fog at severity level 5 images and URIE enhanced images and their confidence scores in the ImageNet-C dataset.
The top and bottom rows are the original and enhanced images, respectively.
}
\label{fig:urie_output_all_010}
\end{figure}

\begin{figure}[t]
\centering
\includegraphics[width=1.0\linewidth]{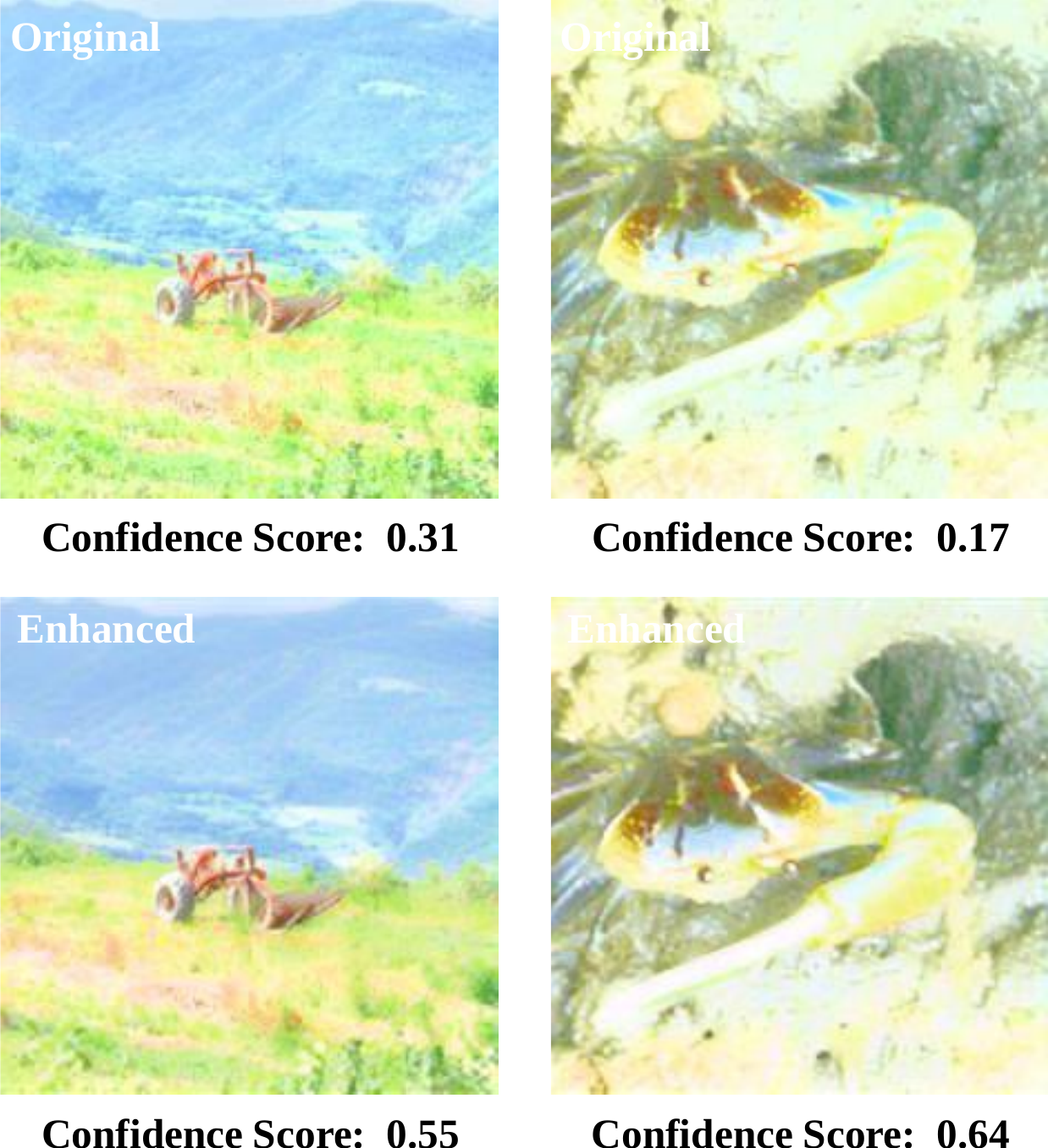}
\caption{
Brightness at severity level 5 images and URIE enhanced images and their confidence scores in the ImageNet-C dataset.
The top and bottom rows are the original and enhanced images, respectively.
}
\label{fig:urie_output_all_011}
\end{figure}

\begin{figure}[t]
\centering
\includegraphics[width=1.0\linewidth]{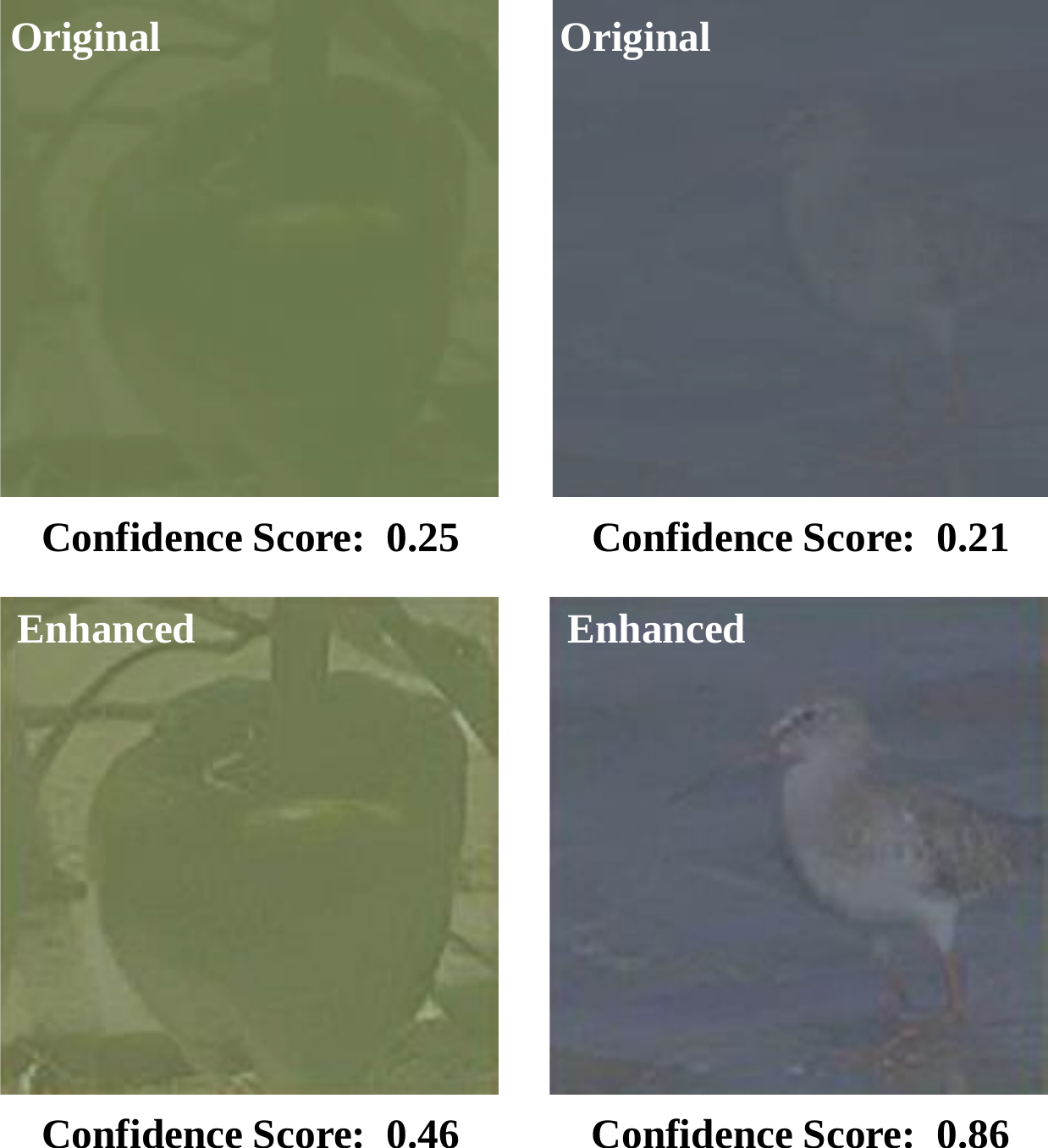}
\caption{
Contrast at severity level 5 images and URIE enhanced images and their confidence scores in the ImageNet-C dataset.
The top and bottom rows are the original and enhanced images, respectively.
}
\label{fig:urie_output_all_012}
\end{figure}

\begin{figure}[t]
\centering
\includegraphics[width=1.0\linewidth]{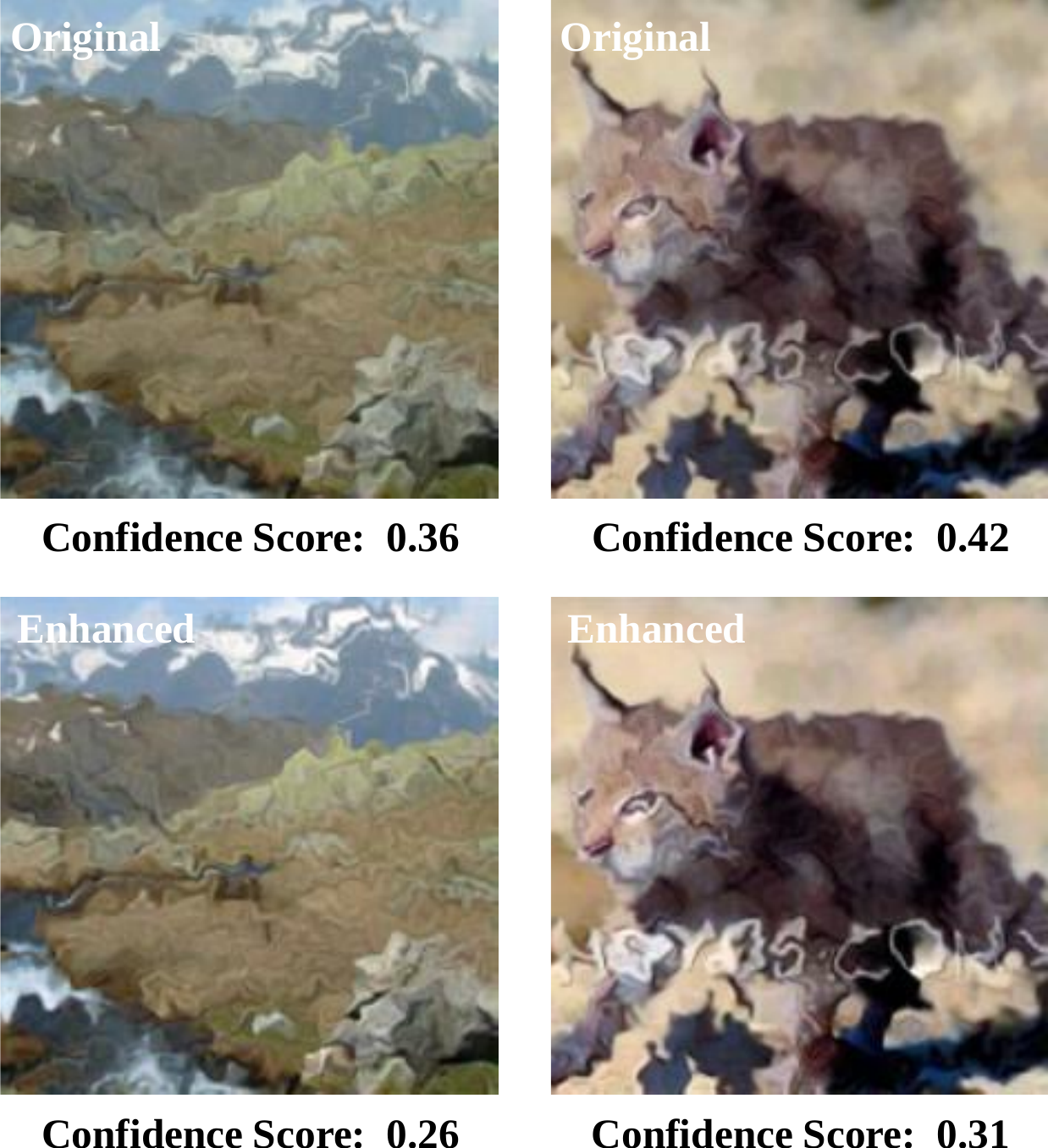}
\caption{
Elastic transform at severity level 5 images and URIE enhanced images and their confidence scores in the ImageNet-C dataset.
The top and bottom rows are the original and enhanced images, respectively.
}
\label{fig:urie_output_all_013}
\end{figure}

\begin{figure}[t]
\centering
\includegraphics[width=1.0\linewidth]{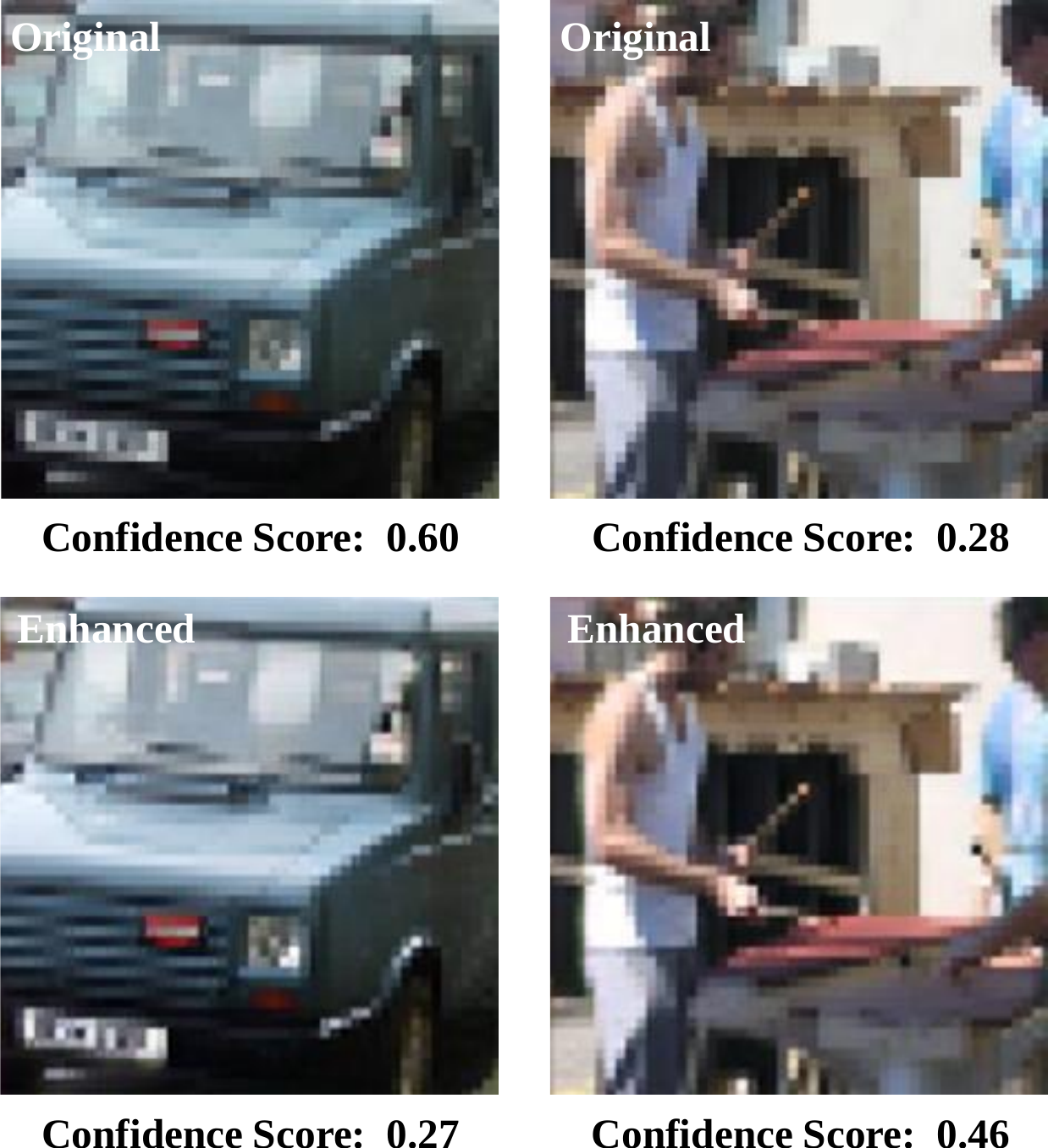}
\caption{
Pixelate at severity level 5 images and URIE enhanced images and their confidence scores in the ImageNet-C dataset.
The top and bottom rows are the original and enhanced images, respectively.
}
\label{fig:urie_output_all_014}
\end{figure}

\begin{figure}[t]
\centering
\includegraphics[width=1.0\linewidth]{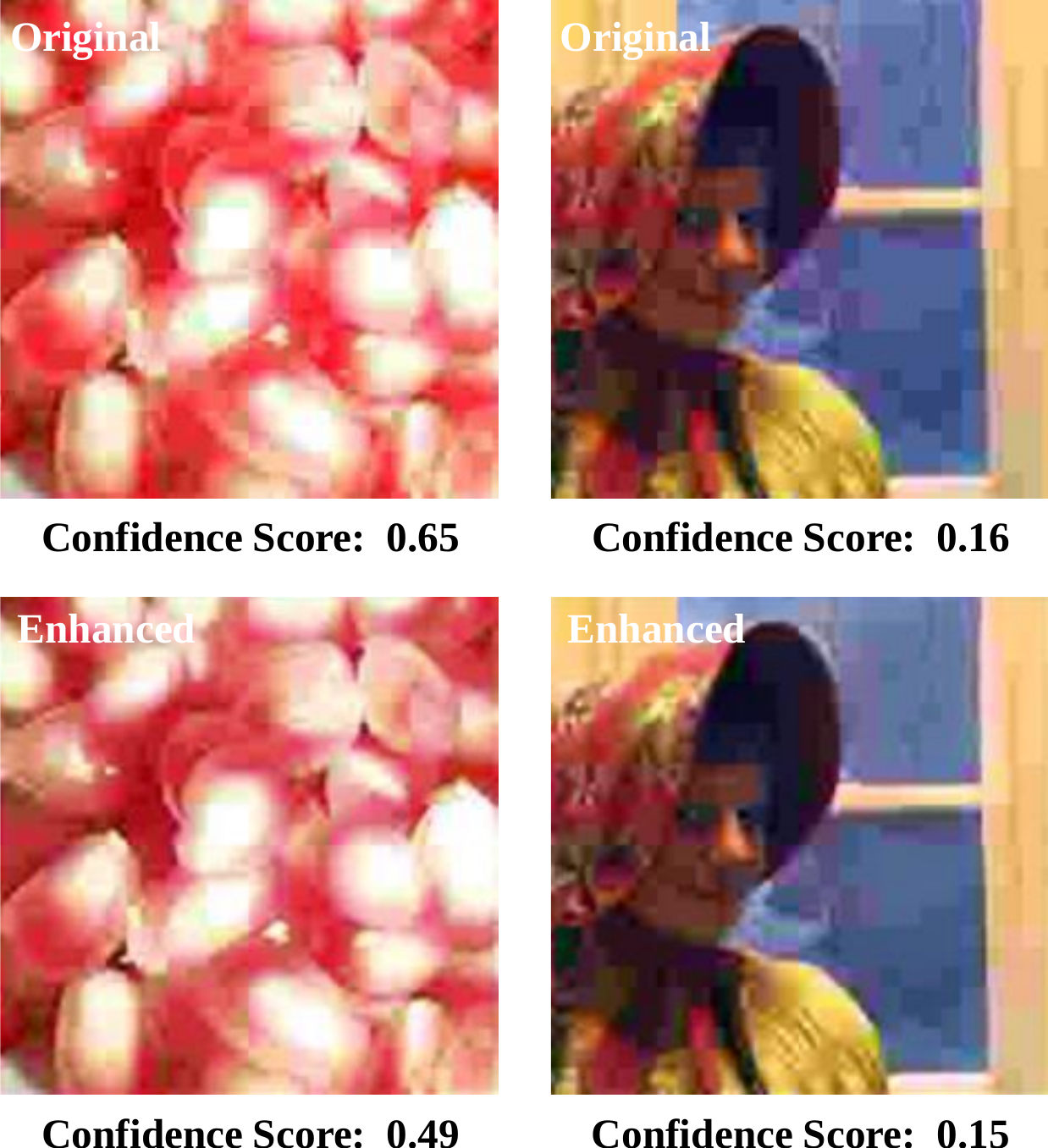}
\caption{
Jpeg compression at severity level 5 images and URIE enhanced images and their confidence scores in the ImageNet-C dataset.
The top and bottom rows are the original and enhanced images, respectively.
}
\label{fig:urie_output_all_015}
\end{figure}

This section provides the same as the visualization in Figure~3, we show the original and enhanced images and their confidence scores for each corruption in the ImageNet-C dataset.
Figure~\ref{fig:urie_output_all_001}--\ref{fig:urie_output_all_015} show the results.

\clearpage

\bibliographystyle{IEEEtran}
\bibliography{egbib}